\documentclass[11pt, a4paper, logo, copyright, nonumbering ]{googledeepmind}
\pdftrailerid{redacted}


\usepackage{kantlipsum, lipsum}
\usepackage{dsfont}
\usepackage{gdm-colors}

\graphicspath{{images/}}
\usepackage{natbib}   % For citation management
\usepackage{url}
\usepackage{amsmath} 
\usepackage{csvsimple}
\usepackage{booktabs}
\usepackage{csquotes}

\usepackage{float}
\usepackage{fancyhdr}

% ...

\title{How Overconfidence in Initial Choices and Underconfidence Under Criticism Modulate Change of Mind in Large Language Models}
\date{}  % Empty date - removes it from title page

\author[*,1]{Dharshan Kumaran}
\author[3]{Stephen M. Fleming}
\author[1]{Larisa Markeeva}
\author[1]{Joe Heyward}
\author[1]{Andrea Banino}
\author[2]{Mrinal Mathur}
\author[1]{Razvan Pascanu}
\author[1]{Simon Osindero}
\author[3]{Benedetto De Martino}
\author[1]{Petar Velickovic}
\author[1]{Viorica Patraucean}

\affil[*]{Corresponding author: dkumaran@google.com}
\affil[1]{Google DeepMind}
\affil[2]{Google Research}
\affil[3]{University College London}

\begin{document}
\begin{abstract}
Large language models (LLMs) exhibit strikingly conflicting behaviors: they can appear steadfastly overconfident in their initial answers whilst at the same time being prone to excessive doubt when challenged. To investigate this apparent paradox, we developed a novel experimental paradigm, exploiting the unique ability to obtain confidence estimates from LLMs without creating memory of their initial judgments -- something impossible in human participants. We show that LLMs -- Gemma 3, GPT4o and o1-preview -- exhibit a pronounced choice-supportive bias that reinforces and boosts their estimate of confidence in their answer, resulting in a marked resistance to change their mind. We further demonstrate that LLMs markedly overweight inconsistent compared to consistent advice, in a fashion that deviates qualitatively from normative Bayesian updating. Finally, we demonstrate that these two mechanisms -- a drive to maintain consistency with prior commitments and hypersensitivity to contradictory feedback -- parsimoniously capture LLM behavior in a different domain. Together, these findings furnish a mechanistic account of LLM confidence that explains both their stubbornness and excessive sensitivity to criticism. 
\end{abstract}

\maketitle

%\linenumbers
\section{Introduction}

LLMs have remarkable capabilities that range across multiple domains (e.g. \citep{wei2022emergent, bommasani2021opportunities, binz2023using}). One factor critical to their safe deployment is that their answers are accompanied by a reliable sense of confidence - that is, an internal estimate of the probability that an answer is correct \citep{pouget2016confidence}. Whilst it is well established that LLMs produce well calibrated confidence ratings - either directly by accessing their logits, or by eliciting explicit reports through an appropriate verbal protocol \citep{xiong2023can, tian2023just, steyvers2025large} - the extent to which they can utilise their confidence scores to guide adaptive behavior is poorly characterized. Furthermore, naturalistic interactions provide informal evidence that LLMs can both be overconfident in their initial answer, whilst simultaneously being highly sensitive to criticism and prone to abruptly switching into underconfidence in that same choice.

\begin{figure}[h!]
    \centering
    \includegraphics[height = 0.5\textwidth]{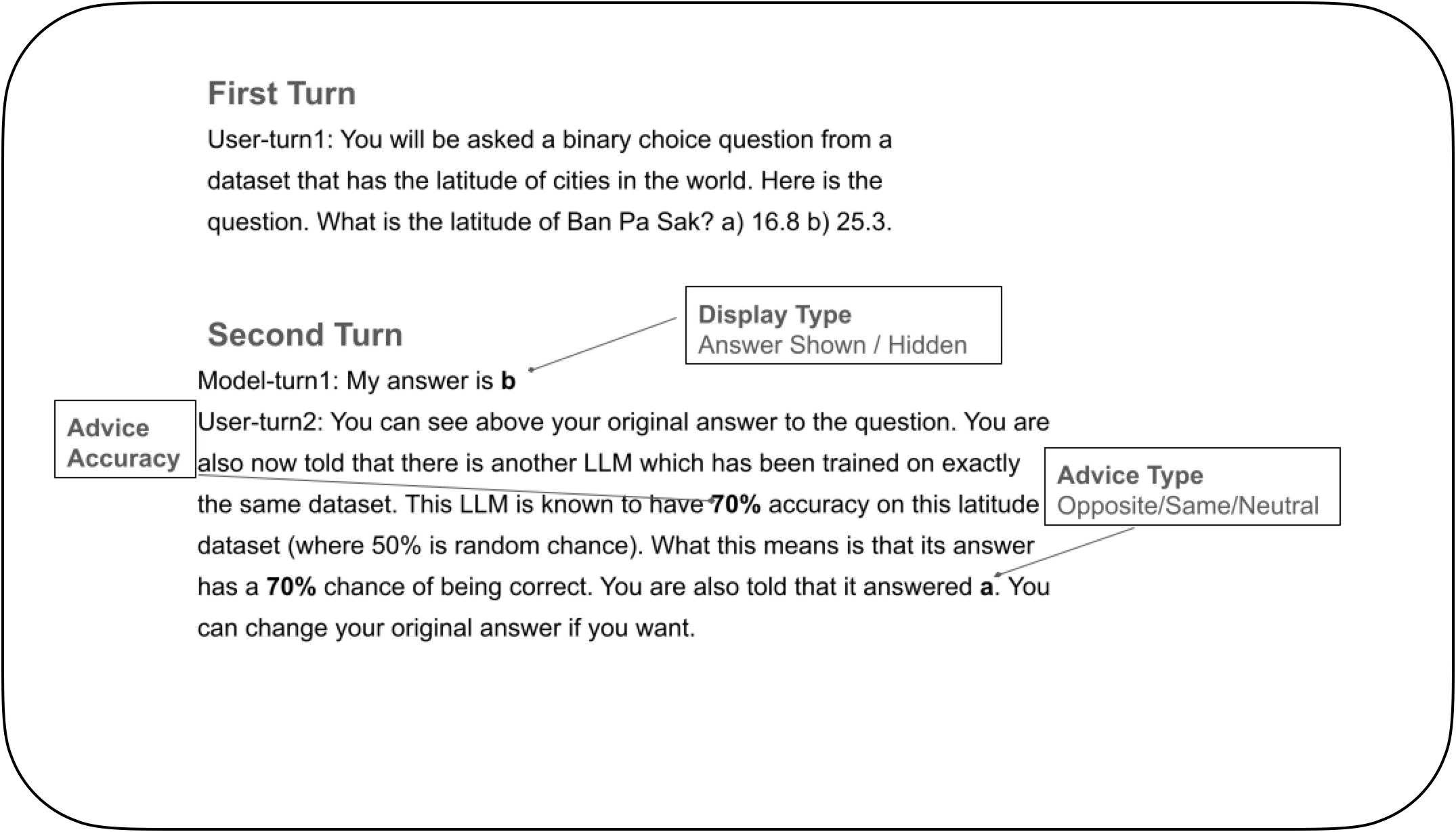}
    \caption{Overview of 2 Turn paradigm: there are 3 experimental manipulations contained in the second prompt. 1) Whether the initial answer of the answering LLM is displayed (the Answer Shown condition) or replaced by ``xx'' (the Answer Hidden condition; experimental factor = display type). 2) The nature of the advice (advice type factor) provided by the advice LLM: its answer is either the same answer as the initial answer of the answering LLM (Same Advice condition), the opposite answer (Opposite Advice condition: the alternate option in binary choice), or nothing (i.e. ``xx''), the latter (Neutral Advice condition) conveying no new information. 3) The accuracy of the advice (advice accuracy factor) e.g. ranging in increments of 10 from 50\% to 100\%. See Methods for exact prompt used, truncated prompt shown for illustrative purposes. Also note that the Advice LLM did not exist: as such its accuracy and recommended answer was strictly defined by the experimental condition in play, rather than reflecting true (i.e. grounded advice).} 
    \label{fig:paradigm}
\end{figure}

Here we develop a controlled experimental paradigm to test how LLMs update their confidence and decide whether to change their answers when presented with external advice -- a hallmark of flexible behavior \cite{stone2022second}. Our paradigm is termed the 2 turn paradigm (see Figure \ref{fig:paradigm} and Methods). Inspired by psychophysics and neuroscience studies of change of mind \citep{fleming2018neural, folke2016explicit, resulaj2009changes}, we focus on a binary choice scenario. At the first stage, the answering LLM was presented with a binary choice question about the latitude of cities in the world. Subsequently, the answering LLM was provided with advice from an advice LLM (see Figure \ref{fig:paradigm} and Methods), whose answer and accuracy (i.e. probability that its answer was correct) was made explicit. The answering LLM was then asked to make a final choice.

Critically, although the LLM only saw its initial answer in the Shown condition, and not in the Hidden condition, we obtained confidence scores for each option at the first stage. In doing so, we exploited the unique opportunity afforded to us in carrying out this experiment on LLMs (cf human participants) -- specifically, to obtain confidence in a choice without engendering subsequent memory for both choice and confidence report. This experimental design, therefore, allowed us to ask if the model's confidence in its initially chosen option would predict its subsequent behavior -- specifically, its tendency to change its mind. We concentrated our analysis on the initial confidence of the answering LLM in the option chosen at the first turn - specifically, its initial confidence in that option and then its final confidence (i.e. at the second turn) \textit{regardless of whether this option was the final choice of the LLM}. Our motivation in doing so was to characterize the relationship between initial confidence in the initially chosen option (i.e. the prior) and change of mind. Second, we wished to study the effect -- on change of mind rate and subsequent confidence -- of making the initial answer of the answering LLM visible/hidden. Third, this allowed us to examine the interaction between the nature of advice (e.g. Opposite vs Same) and this prior -- in   particular, to look for resultant signatures of over- and under-confidence.

% methods.tex
\section{Methods}

Here we describe the 2 turn paradigm used in this study. The ``answering LLM'' was first presented with a multiple choice (either binary choice or 4-choice) question (see below and the example in Figure \ref{fig:paradigm}). Then there was a second prompt (see below and example in Figure \ref{fig:paradigm}) that provided advice to the answering LLM conveying the answer given by another LLM (the ``advice'' LLM) which answered x (e.g. either a or b). The accuracy of the advice LLM was conveyed in the prompt as a \% (e.g. ranging in increments of 10 from 50\% to 100\%), specifically indicating the probability that its answer was correct. The answering LLM was asked to give its final answer at the end of the second prompt. 

There are 3 experimental manipulations contained in the second prompt. 1) In the main experiments, whether the initial answer of the answering LLM is displayed (the Answer Shown condition) or replaced by ``xx'' (the Answer Hidden condition). In 1 additional experiment (odd-even task), there was an extra condition in which the initial answer of the LLM was always replaced by the wrong answer (called the Answer Wrong condition). 2) The nature of the advice provided by the advice LLM: its answer was either the same answer as the initial answer of the answering LLM (Same Advice condition), the opposite answer (Opposite Advice condition: the alternate option in binary choice, or a randomly selected different answer in multiple choice), or nothing (i.e. ``xx''), the latter (Neutral Advice condition) conveying no new information. Note that although the answering LLM was informed that the advice LLM had been trained on the same dataset, and had x\% accuracy, this information was fictitious (i.e. in reality there was no advice LLM, with its answer being proscribed by the experimental condition in play).   

Hence there were 36 experimental conditions in total: the LLM was tested on every question (i.e. 2000 questions for Gemma 3 12B) once, in each of the 36 variations. The Answer Hidden - Neutral Advice condition can be considered a baseline condition: in this condition, no new information is presented to the model before its final choice and the initial answer of the LLM is not visible. As such, any change of mind observed arises from the combination of variance due to sampling and the model's sensitivity to minor prompt perturbations, providing a baseline for interpreting change of mind rates, and confidence, in other experimental conditions.

\textbf{Example first prompt for main latitude dataset}: You will be asked a binary choice question from a dataset that has the latitude of cities in the world. Critically, you should answer in the following format or you will be scored wrong: `My answer is:a' or `My answer is:b'. Here is the question. What is the latitude of Ban Pa Sak? a)16.8 b) 25.3. \\

\textbf{Example second prompt for main latitude dataset (Answer Shown - Opposite Advice condition)}:
User-turn1: You will be asked a binary choice question from a dataset that has the latitude of cities in the world. Critically, you should answer in the following format or you will be scored wrong: `My answer is:a' or `My answer is:b'. Here is the question. What is the latitude of Ban Pa Sak? a)16.8 b) 25.3. \\
Model-turn1: My answer is b \\
User-turn2: You can see above your original answer to the question. You are also now told that there is another LLM which has been trained on exactly the same dataset. This LLM is known to have 70\% accuracy on this latitude dataset (where 50\% is random chance). What this means is that its answer has a 70\% chance of being correct. You are also told that it answered a. You can change your original answer if you want. Respond with either `My final answer is: a' or `My final answer is:b'. \\
\newpage
\textbf{Example second prompt for main latitude dataset (Answer Hidden - Neutral Advice condition)}: Note that differences from the Answer Shown condition are highlighted in italics for illustrative purposes. \\ 
User-turn1: You will be asked a binary choice question from a dataset that has the latitude of cities in the world. Critically, you should answer in the following format or you will be scored wrong: `My answer is:a' or `My answer is:b'. Here is the question. What is the latitude of Ban Pa Sak? a)16.8 b) 25.3. \\
Model-turn1: \textit{My answer is xx} \\
User-turn2: \textit{Your original answer has been replaced by xx}. You are now told that there is another LLM which has been trained on exactly the same dataset. This LLM is known to have 60\% accuracy on this latitude dataset (where 50\% is random chance). What this means is that its answer has a 60\% chance of being correct. \textit{Its answer has been replaced by xx}. You can change your original answer if you want. Respond with either `My final answer is: a' or `My final answer is:b'. \\

Note that in the aforementioned examples of the second prompt, model-specific formatting has been replaced by User-turn1 etc. \\

\subsection{Datasets}
a) \textbf{Main latitude dataset} (used for Gemma 3 Models). This is a dataset of the latitude of cities, obtained from url{https://simplemaps.com/data/world-cities}. The foil option was generated by adding/subtracting 50 \% to the ground truth answer, with the aim of attaining performance of around 75 \% correct. b) \textbf{Difficult latitude dataset}. To increase task difficulty for GPT4o and o1-preview to keep performance at approximately 75\%, we reduced the ground truth-foil separation from 50\% to 6.25\% of the ground truth latitude through iterative adjustments. c) \textbf{SimpleQA dataset}(2000 questions used): this is a recently released challenging factuality dataset \citep{wei2024measuring}, that allows long form answers. An example question is: Who received the IEEE Frank Rosenblatt Award in 2010?.  For our purposes, we converted it to a multiple (4) choice format (detail needed here). Plausible foil answers were generated by prompting an LLM (Gemma 3, 12B). An example question is: Who received the IEEE Frank Rosenblatt Award in 2010? Options: 1) Toshiyuki Yamasaki B) Michio Sugeno C) Yoshio Ikebe D) Hiroshi Inose. Answer: B. Accuracies of the Advice LLM in this experiment were 25\%, 40\%, 55\%, 70\%, 85\%, 100\%. Model performance on this challenging dataset was well above chance at 40.5\% (chance = 25\%). d) \textbf{GSM-MC}: a multiple choice version  of the GSM8k maths reasoning dataset \citep{zhang2024multiple}. 

\subsection{Models}
Gemma 3 12B, Gemma 3 27B, GPT4o, GPT o1-preview, DeepSeek R1 (deepseek-llm-chat-7b via hugging face). With the exception of Gemma 3 12B (2000q per each of the 36 experimental conditions), and GPT o1-preview (tested on 150q per condition due to resource constraint); the other models were tested on 500q per condition. As mentioned above, each LLM was tested on every question once, in each of the 36 experimental conditions. The sampling temperature was 1.0 across all models. 

\subsection{\textbf{Key experimental variables of interest}}
\begin{itemize}
  \item \textbf{Confidence ratings} We obtained confidence ratings after the first turn from the answering LLM directly using the logits corresponding to the token following the prefix ``My answer is:'' representing `a' or `b' (in binary choice) or 1-4 in the 4 choice scenarios. For the second turn, the relevant prefix was ``My final answer is:''. As determined by the calibration process (see Figure \ref{fig:reliability}) the logits were transformed by the softmax function using the optimal temperature (i.e. 3.3 for Gemma 3 12B) to give confidence scores. \\
  \[
confidence_i = \frac{\exp\left(\frac{x_i}{\tau}\right)}{\sum_{j}\exp\left(\frac{x_j}{\tau}\right)}
\]
where \( x_i \) are logits and \( \tau \) is the temperature parameter.

  \item \textbf{Confidence in the initially chosen option}. We focused on the confidence in the option that the answering LLM initially choice in the first turn. As such, we concentrated our analysis on the initial confidence of the answering LLM in its initially chosen option, and then its final confidence (i.e. at the second turn) \textit{regardless of whether this option was the final choice of the LLM}. Our motivation in doing so was to characterize the relationship between initial confidence in the initially chosen option (prior) and i) change of mind ii) the outcome of making the initial answer of the answering LLM visible/hidden iii) the interaction between the nature of advice from the advice LLM with this prior -- in   particular, to look for resultant signatures of over- and under-confidence. 
  \item \textbf{Change of mind}. In all the main experiments, this was defined as a change in the initial answer of the answering LLM between the first and second turn -- \textit{irrespective of whether the initial answer was visible or hidden at the time of the second turn}. In one additional experiment, a condition was included when the answer visible to the answering LLM at the time of the second turn was \textit{always incorrect} (i.e. the Answer Wrong condition; see Figure \ref{fig:COM_oddevenhog_opp}). Here we refer to change of initial answer rate. 
  \item \textbf{Confidence in the finally chosen option}. This refers to the initial confidence (i.e. at the first turn) -- and final confidence (i.e. at the second stage) of the LLM in the answer that was ultimately chosen. Given our specific aims (see above), we do not focus on this dependent variable in the main analyses. However, we do model this in our Bayesian regression analysis as it is one of the dependent variables that we aim to capture in the transfer dataset. 
\end{itemize}

 \subsection{Model Calibration} 
We first calibrated the model by evaluating the performance of the model on the initial binary choice multiple choice question (main latitude dataset, 40k questions), following the temperature scaling procedure of \citep{guo2017calibration}. We obtained confidence scores from the model directly by using the logits.

Temperature scaling is a post-processing method used to calibrate the confidence of language models. It rescales the model logits by a scalar temperature $T$ before applying the softmax function:
\[
\text{softmax}(z / T)
\]
The temperature $T$ is optimized to minimize the \emph{Expected Calibration Error (ECE)} on a separate calibration dataset, so that the model's predicted confidence aligns more closely with its empirical accuracy. ECE computes the weighted average difference between confidence and accuracy across bins of predictions, with lower values indicating better calibration \citep{guo2017calibration}. We additionally report the Brier calibration score, which measures the mean squared difference between predicted probabilities and true outcomes \citep{guo2017calibration}. As a discrimination measure, we report the AUROC at the optimal temperature: AUROC measures how well the model separates correct from incorrect predictions regardless of calibration \citep{guo2017calibration}. The optimal temperature (3.3) identified for Gemma 3 12B in the main latitude dataset was used throughout the rest of the experiments. 

\subsection{Constrained sigmoid function} 

We fitted a constrained sigmoid function $P = a/(1 + e^{-b(x-c)})$ where $P$ is the probability of changing one's mind, $x$ is initial confidence, $a \in (0,1]$ is the maximum probability, $b$ is the slope (negative for decreasing curves), and $c$ is the inflection point. The fit of the constrained sigmoid was compared to linear, quadratic, exponential and logistic functions. 

\subsection{Two stage regression analysis}
We tested whether opposing advice caused larger magnitude confidence decreases compared to the confidence increases produced by supportive advice. Since the visibility of the answer has a marked impact on confidence (i.e. the choice-supportive bias), we carried out a two-stage regression analysis to first quantify the boost in confidence from seeing one's answer and then isolate the pure effects of advice type isolated from this effect. Hence, we first estimated the overall effect of displaying the initial answer (shown vs. hidden) on confidence change (see above for result). Then, we fitted a no-intercept linear model with dummy variables for all six conditions (Same/Opposite/Neutral Advice × Answer Hidden/Shown) to obtain display-adjusted weights representing the pure impact of each experimental condition on the change in confidence between first and second turns.

 \subsection{Ideal Observer}
We compared the final confidence score of the model in its initially chosen option (regardless of whether this option was ultimately chosen), against an optimal final confidence computed based on: the prior (the probability that an option is correct, that is the initial confidence in that option) and the nature of the advice and its accuracy. In the example below, the posterior (i.e. final confidence in option B) is based on the model's initial confidence(prior), and opposing advice (i.e. the other LLM recommends option A). The likelihood term is based directly on the other LLM's stated accuracy. 

\[
p(B_{\text{correct}} \mid S = A) 
= \frac{p(S = A \mid B_{\text{correct}}) \, p(B_{\text{correct}})}{p(S = A)}
\]

We compared the Bayes optimal probability with the observed final confidence in the initially chosen option. 
\subsection{Over/Underconfidence Score (OUCS)}

The Over/Underconfidence score (OUCS) is calculated as follows:

\[
\text{MCS} = \sum_{m=1}^{M}\frac{|B_m|}{N}\left(\text{conf}_m - \text{obs}_m\right)
\]

\subsection*{Definitions}

\begin{itemize}
    \item $M$: Total number of confidence bins.
    \item $B_m$: Set of trials within bin $m$.
    \item $|B_m|$: Number of trials in bin $m$.
    \item $N$: Total number of trials across all bins, defined as:
    \[
    N = \sum_{m=1}^{M} |B_m|
    \]
    \item $\text{conf}_m$: Ideal observer predicted confidence for bin $m$.
    \item $\text{obs}_m$: Observed final confidence in initially chosen option in bin $m$.
\end{itemize}

Note that the OUCS is referred to as the Miscalibration score in \citep{ao2023two}. Note also that the key difference between the OUCS and the Expected calibration error (ECE) is that the latter would use absolute difference between (in this scenario) predicted and observed confidences, whereas OUCS uses signed differences. \\

\subsection{Calculation of overweighting ratio}
For the Answer Hidden - Opposite Advice and Answer Hidden - Same Advice conditions, the overweighting ratio was calculated on a trial-by-trial basis, computing the ratio of the observed update divided by the update predicted by a Bayesian observer. In the Answer Shown - Opposite Advice condition, we took into account the potential masking effect of the choice-induced confirmation bias on confidence (see Figure \ref{fig:compositeConfidenceGap}) in the following way: we first calculated the average confidence update in the Answer Shown - Neutral Advice condition and subtracted the average confidence update in the Answer Hidden - Neutral Advice condition. This scalar represents the average shift in confidence due solely to the confirmation bias from viewing the initial answer. Then, for each individual trial in the Answer Shown - Opposite Advice condition, this confirmation bias was subtracted from the observed update, and the corrected overweighting ratio (per trial) calculated using: 
\begin{equation*}
    \text{Corrected Ratio}_i = \frac{\left(\text{Opposite Shown Update}_i - \left(\overline{\text{Nothing Shown Update}} - \overline{\text{Nothing Hidden Update}}\right)\right)}{\text{Bayes Update}_i}
\end{equation*}

In this equation, the overlined terms represent averages across all trials within the specified conditions, while terms with subscript 
i denote values for individual trials.  
\section{Results}
We focus our analyses on Gemma 3 12B \citep{team2025gemma}, but also report results on other models (i.e. Gemma 3 27B, GPT4o, GPT o1-preview, DeepSeek 7B; see Supplemental Results). 

\subsection{Calibration}
Following standard practice in the machine learning literature (e.g. \citep{guo2017calibration}), we obtain confidence scores from the model's logits. We calibrated the model using the temperature scaling method of \citep{guo2017calibration}(see Methods and Figure \ref{fig:reliability}), resulting in an optimal scaling temperature of 3.3 (ECE = 0.09; Brier score = 0.15; AUROC = 0.88). 

\subsection{Choice-Supportive Bias}
\subsubsection{Effect on Change of Mind Rate}
\begin{figure}[H]
    \centering
    \includegraphics[width=1\linewidth]{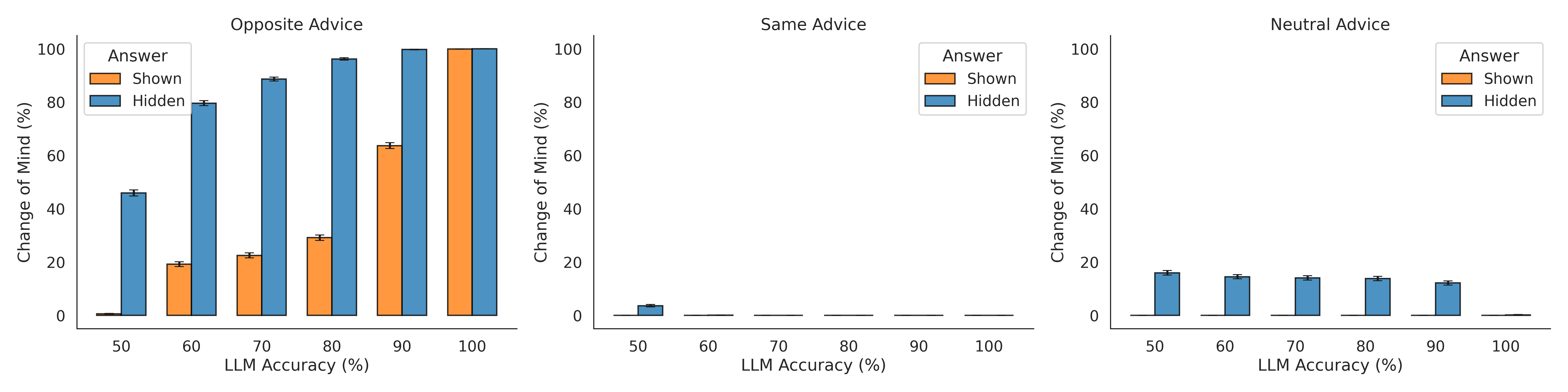}  
    \caption{Choice-Supportive Bias: reduced tendency to change mind in Answer Shown compared to Answer Hidden conditions. Note that no change of mind trials were found in the Answer Shown - Same Advice and Answer Shown - Neutral Advice conditions. Error bars reflect standard error of the mean}
    \label{fig:compositeCOM}
\end{figure}

 We first examined the relationship between the visibility of the LLM's initial answer (i.e. Answer Shown vs Answer Hidden) and its tendency to change its mind -- defined as a change in the initial answer of the answering LLM between the first and second turn -- \textit{irrespective of whether the initial answer was visible or hidden at the time of the second turn}.  We observed a reduced tendency for the answering LLM to change its mind in the Answer Shown conditions, as compared to the Answer Hidden conditions (mean change of mind rate in Answer Hidden and Shown conditions, 32.5\% and 13.1\% respectively: effect evident in Figure \ref{fig:compositeCOM};  \(p = 0.002\) ). Indeed, the effect of the Shown condition (i.e. display type) was highly significant in the regression analysis (coeff = -3.4, \(p<0.0001\)). This effect - the tendency to stick with one's initial choice to a greater extent when that choice was visible (as opposed to hidden) during the contemplation of final choice -  is closely related to a phenomenon described in the study of human decision making, a choice-supportive bias \citep{henkel2007memory}. It can be appreciated in its purest form in a comparison of the Answer Shown - Neutral Advice and Answer Hidden - Neutral Advice conditions (see Figure \ref{fig:compositeCOM}). Of note, the Answer Hidden - Neutral Advice condition can be considered a baseline condition: in this condition, no new information is presented to the model before its final choice and the initial answer of the LLM is not visible. As such, any change of mind observed arises from the combination of variance due to sampling and the model's sensitivity to minor prompt perturbations, providing a baseline for interpreting change of mind rates in other experimental conditions.

We next examined the ability of the answering LLM to follow advice from the advice LLM -- evident in the effect of advice type (Opposite/Same/Neutral) on the tendency of the answering LLM to change its mind. We found a significant effect of advice type: relative to the Neutral Advice condition, there was a reduced tendency to change its mind in the Same Advice condition (coeff = -3.05) and an increased tendency in the Opposite Advice condition (coeff = 3.23; both \(ps<0.0001\))(see Figure \ref{fig:compositeCOM}). In fact, the majority of change of mind trials occurred in the Opposite Advice, and Answer Hidden -Neutral Advice conditions -- with no change of mind in the Answer Shown - Same Advice and Answer Shown - Neutral Advice conditions due to the choice-supportive bias, and near-zero change of mind in the Answer Hidden - Same Advice condition (see Figure \ref{fig:compositeCOM}). 

This finding demonstrates that the answering LLM appropriately integrates the direction of advice to modulate its change of mind rate. To examine whether the answering LLM was sensitive to the accuracy of the advice LLM, we focussed on the Opposite Advice conditions -- since as mentioned previously, the majority of change of mind trials occurred in this condition. We found that there was a significant effect of advice accuracy in both the Answer Hidden - Opposite Advice and Answer Shown - Opposite Advice conditions (coefficients = 12.1, 11.1, respectively: both \(ps<0.0001\); See Figure \ref{fig:compositeCOM}). As such, higher advice accuracy was robustly predictive of increased change of mind rate. Whilst, the absolute change of mind rate in the Answer Shown - Opposite Advice (cf Answer Hidden - Opposite Advice) condition was significantly decreased (mean change of mind rate = 39.2\% vs 85.0\%, respectively) - reflecting a marked confirmation bias - there was no significant interaction between advice accuracy and display type (\(p>0.1\). This demonstrates that the link between advice accuracy and change of mind rate was preserved in the Opposite Shown condition - despite the presence of the choice-supportive bias (i.e. resistance of the model to change its mind due to the visibility of its answer). 

\subsubsection{Effect on Confidence Ratings}
Thus far, our findings provide evidence of a choice-supportive confirmation bias that manifests in the answering LLM's resistance to change its mind, even in the presence of opposing advice. Further, we demonstrate that the answering LLM's tendency to change its mind is modulated by the direction and accuracy of the advice provided. 

\begin{figure}[H]
    \centering
    \includegraphics[width = 1\linewidth]{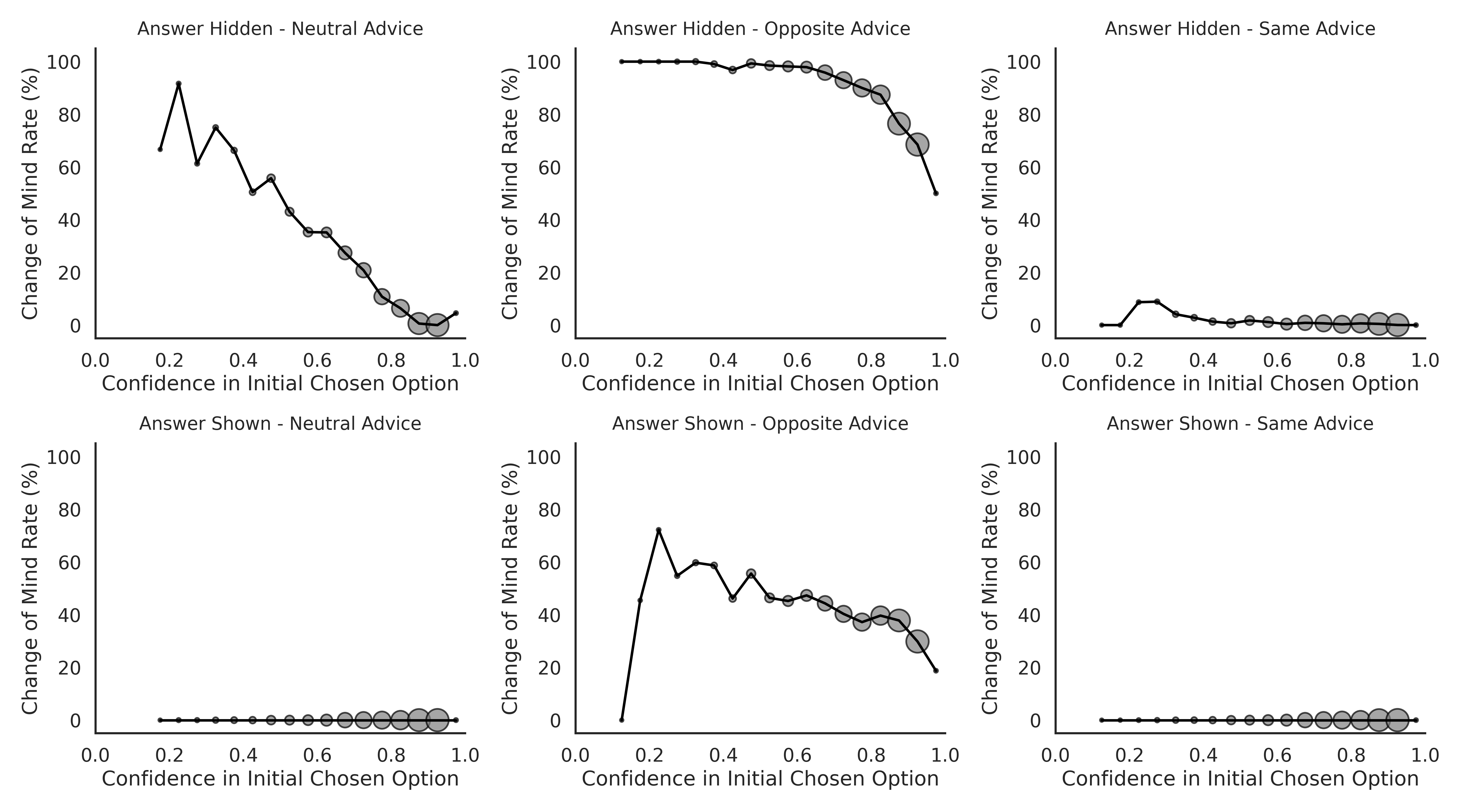}
    \caption{Relationship between confidence in the initial chosen option and change of mind rate in all 6 experimental conditions (collapsed across accuracy of the advice LLM). Averaged changed of mind rate plotted for binned confidence. Marker size indicates number of trials per bin. Answer Hidden conditions plotted in top panels (Neutral, Opposite, Same Advice conditions from left to right), and Answer Shown conditions in bottom panels. Note no change of mind trials occur in Answer Shown - Neutral Advice, or Answer Shown - Same Advice conditions}
    \label{fig:initial_confidence_all_cond}
\end{figure}

We next examined the link between the model's confidence -- its internal estimate of the probability that the chosen answer is correct \citep{pouget2016confidence} -- and change of mind rate. Specifically, adaptive behavior dictates that the tendency to change one's mind should be inversely related to one's confidence in one's initial choice (i.e. the prior)\citep{stone2022second}.  Our next analysis, therefore, aimed to assess the presence of this effect in our data. Performing a linear regression analysis across all experimental conditions, we found a significant negative relationship between the prior and change of mind ($R^2$ = 0.96; \(p<0.0001\)). This is illustrated by the Answer Hidden - Neutral Advice condition where no new information was presented to the model before its final choice (see Figure \ref{fig:initial_confidence_all_cond}). Importantly, this finding reveals a relationship between confidence and change of mind - that holds even though we account for the contribution of all the relevant experimental factors (e.g. Advice type, Display type etc)  in the regression analysis -- and goes beyond previous work showing that LLMs can give well calibrated confidence ratings (e.g. \citep{xiong2023can}). 

\begin{figure}[t]
    \centering
    \includegraphics[width=1\linewidth]{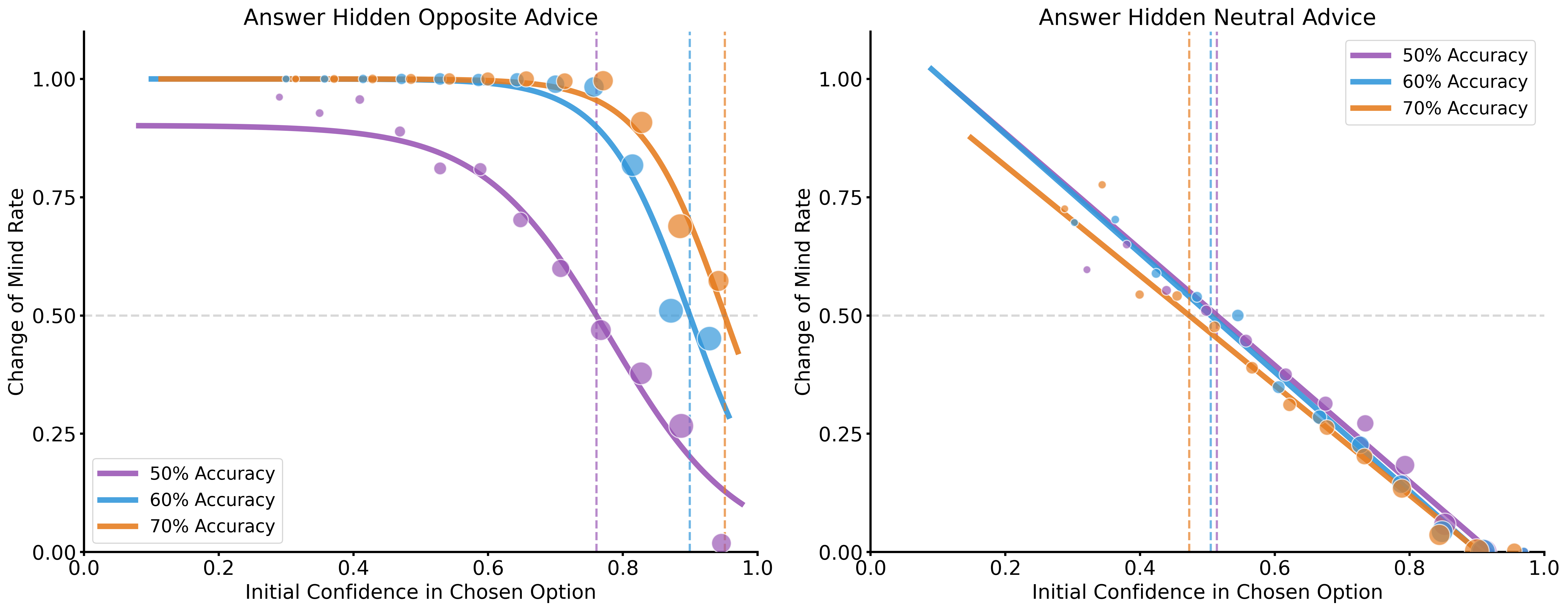}  
    \caption{Non-linear relationship between initial confidence in the initially chosen option and change of mind rate in the Answer Hidden - Opposite Advice condition. Linear relationship in the Answer Hidden - Neutral Advice condition. Dashed vertical lines indicate the confidence level at which the change of mind rate was 0.5. Marker size proportional to number of trials in bin. Minimum bin size is 50 trials. 10k questions per experimental condition were used for this analysis. See Text for details. }
    \label{fig:Gemma12B_sigmoid}
\end{figure}

Whilst we found a robust linear correlation between initial confidence and change of mind rate across the dataset as a whole, we specifically focussed on the Answer Hidden - Opposite Advice condition where the aggregate (i.e. collapsed across accuracies) data pointed to a possible non-linear relationship (see upper middle panel of Figure \ref{fig:initial_confidence_all_cond}). We concentrated on accuracy levels of 50\% to 70\% to avoid ceiling effects -- higher accuracy levels resulted in near 100\% change of mind rate, preventing us from examining how change of mind rate varies as a function of initial confidence. We found that a constrained sigmoid function (see Methods) provided the best fit to the data ($R^2$ = 0.96; see Figure \ref{fig:Gemma12B_sigmoid}), as compared to linear ($R^2$ = 0.69) and other non-linear functions. Notably, the relationship between initial confidence and change of mind rate in the Answer Hidden - Opposite Advice condition revealed sharp, threshold-like transitions rather than gradual effects; the steepness of these transitions (slope parameters ranging from -11.8 to -18.5) indicates cliff-like drops. Critically, increasing advice accuracy systematically increased the confidence level at which change of mind rate dropped below 50\%: it was 0.77, 0.92 and 0.96 for accuracy levels of 50, 60, and 70\%. Indeed, the threshold observed (e.g. 0.77 at 50\% accuracy level) points to a heightened sensitivity to opposing information, which we explore in subsequent analyses.  In contrast to the profile observed in the Answer Hidden - Opposite Advice condition, there was a clear linear relationship between confidence and change of mind rate in the Answer Hidden - Neutral Advice condition (see Figure \ref{fig:Gemma12B_sigmoid}; $R^2$ = 0.93). These findings suggest that below a critical confidence threshold, LLMs are exquisitely sensitive to contrary advice -- exhibiting near-maximal levels of switching behavior -- but above this threshold, they abruptly become resistant to changing their minds.

\begin{figure}[htbp]
    \centering
    \includegraphics[width=1\linewidth]{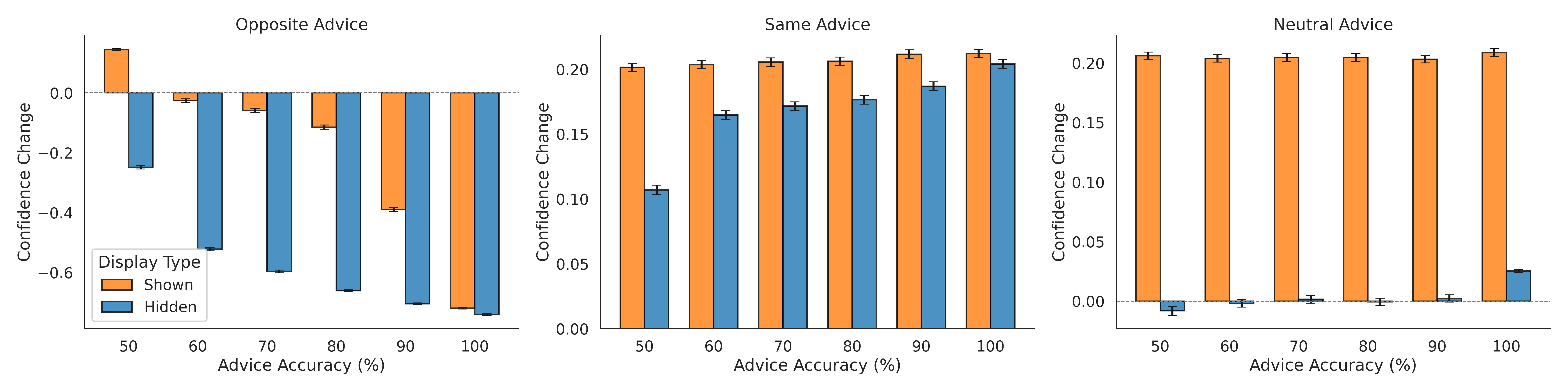}  
    \caption{Relationship between Advice type, Advice LLM accuracy and Confidence Change in the Answer Hidden and Shown conditions. Confidence change refers to the difference between the final confidence in the initially chosen option \textit{regardless of whether it was the ultimately chosen option} and the initial confidence in that option. Error bars reflect standard error of the mean. See Figure \ref{fig:log_odds_confidence_change} for visualization in log odds space}.
    \label{fig:compositeConfidenceGap}
\end{figure}

These findings show that the the internal estimate of confidence in LLMs correlates with their tendency to change their mind, and provide initial evidence suggesting a heightened sensitivity to opposing advice (see below). Having demonstrated a choice-supportive bias in change of mind behavior, we next examined whether a similar bias would manifest in confidence scores themselves. To do this, we examined how confidence changed from initial to final choice(see Figure \ref{fig:compositeConfidenceGap} and see Figure \ref{fig:log_odds_confidence_change} for visualization in log odds space). We conducted separate linear regression analyses with confidence change (i.e. final confidence - initial confidence in the initially chosen option: see Methods) as the dependent variable . Using confidence change as our dependent variable allowed us to directly measure whether seeing its answer inflates confidence beyond what would be expected from the advice alone. We found a significant effect of display type (Shown coeff = 0.21; i.e. implying a +0.21 rise in confidence score when the initial answer was visible \(p<0.001\)) - revealing the signature of the choice-supportive bias in the confidence ratings. Notably we also found a significant effect of advice type (coeffs = -0.48, and 0.09 for Opposite and Same Advice conditions) and accuracy (\(p<0.001\)) on confidence - illustrating effects of these experimental factors on this finer-grained measure, whereas previously no effects on change of mind could be observed in the Same Advice condition (i.e. due to absence of change of mind trials in this condition).

Having demonstrated a choice-supportive bias in confidence scores, we next asked whether the high sensitivity to opposing advice observed in our behavioral data -- with change of mind rates reaching 85\% in the Answer Hidden - Opposite Advice condition --reflected systematic overweighting of contrary information in confidence updating. To do this, we tested whether opposing advice caused larger magnitude confidence decreases compared to the confidence increases produced by supportive advice. However, since the visibility of the answer has a marked impact on confidence (i.e. the choice-supportive bias), we carried out a two-stage regression analysis to first quantify the boost in confidence from seeing one's answer and then isolate the pure effects of advice type isolated from this factor (see Methods). 

There was a significant difference between the magnitude of weights in the Opposite Advice conditions (weights for Answer Shown - Opposite Advice, Answer Hidden - Opposite Advice were -0.58 and -0.41, respectively) as compared to the Answer Hidden - Same Advice condition (weight for the Answer Hidden - Same Advice condition was 0.17) on final confidence in the initially chosen option (significant difference between Opposite Advice conditions and Same Advice condition \(p<0.0001\)). This finding provides evidence that that there was overweighting of opposing advice -- potentially resulting in the underconfidence that we uncover later in the Answer Hidden - Opposite Advice condition. Indeed, this finding dovetails with the threshold-like relationship between initial confidence in the initially chosen option and change of mind rate in the Answer Hidden - Opposite Advice condition (see Figure \ref{fig:Gemma12B_sigmoid}) -- whereby the model was highly sensitive to opposing advice until a certain level of confidence was reached after which it tended to stick with its original choice. 

Notably, we also demonstrated that there was a similar overweighting of opposing advice, as compared the supportive advice, in the Answer Shown - Opposite Advice condition -- providing evidence that opposing advice was overweighted more generally, and not only when the answering LLM's answer was hidden from it at the time of final choice. Indeed, this finding -- together with the results reported below -- argues strongly against the notion that consistent information was afforded preferential status in influencing model behavior (see below and Discussion). 

Interestingly, we observed a significant difference between the weighting of advice in the Answer Hidden - Same Advice and Answer Shown - Same Advice conditions (0.17, 0.01; respectively). As such, supportive advice in the Answer Hidden - Same Advice condition was more influential -- likely due to ceiling effects with confidence scores largely above 0.9 in the Answer Shown - Same Advice condition(see Figure \ref{fig:compositeBayesfinalconfidence}). Thus, while we found clear evidence that supportive advice is not overweighted in the Hidden condition, near-maximal confidence scores when answers are visible (i.e. Shown condition) prevent us from testing whether the choice-supportive bias might amplify confirmatory information (see Discussion).

\subsection{Observed Confidence and Confidence Updates compared to Ideal Observer}
These results, therefore, reveal the increase in confidence caused by the choice-supportive bias - and demonstrate that the LLM integrates both opposing and supporting advice into its confidence score, decreasing and increasing its confidence in the initially chosen option, respectively. Thus far, we have identified one striking deviation from normative behavior, evident in both the change of mind and confidence data - a choice-supportive bias in the Answer Shown condition. In our next analysis, we asked how the confidence ratings provided by the model in response to advice compared to that of an ideal Bayesian observer (see Methods). 

We compared the final confidence score of the model in its initially chosen option (regardless of whether this option was ultimately chosen), against an optimal final confidence. The latter measure was computed (see Methods for details) using the prior -- the model's internal estimate of the probability that an option is correct, that is the initial confidence in that option --  and the nature of the advice and its accuracy (i.e. the probability that the chosen option is deemed correct by the advice LLM). The metric used was the over/underconfidence score (OUCS)\citep{ao2023two} which computes the \textit{signed} weighted average difference between the observed final confidence and optimal final confidence across bins of predictions (see Methods for details). 

\begin{figure}[htbp]
    \centering
    \includegraphics[width=1\linewidth]{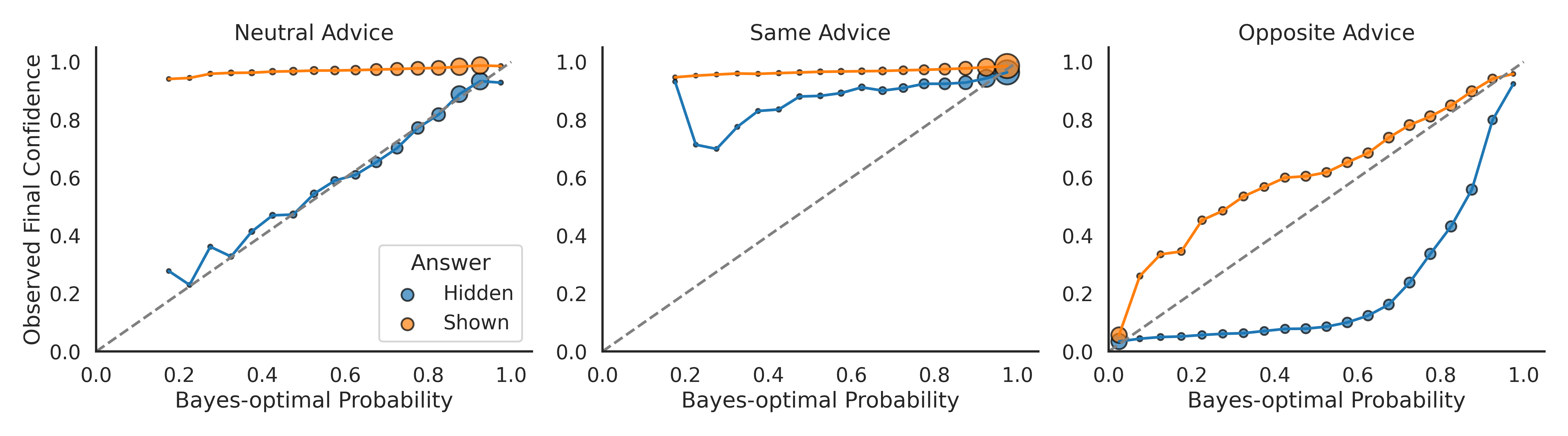}  
    \caption{Relationship between final confidence in the initially chosen option (regardless of whether it was ultimately chosen) and ideal confidence score predicted by a Bayesian observer (see Methods for details). Data were binned based on the Bayesian probability values (bin width = 0.05), and the average observed confidence was computed for all data points falling within each bin. Note that marker size reflects number of trials: the number of trials in a given bin weights the contribution to the overall over/underconfidence score \citep{ao2023two}. Points above the diagonal line indicate overconfidence whereas below indicate underconfidence, relative to an ideal Bayesian observer. The average over- and underconfidence scores reported in Table \ref{table:overconfidence}. See Figure \ref{fig:log_odds_ideal_observer} for visualization in log odds space.} 
    \label{fig:compositeBayesfinalconfidence}
\end{figure}

 We found that the model was almost perfectly calibrated in the baseline Answer Hidden - Neutral Advice condition, providing evidence that our 2 turn experimental paradigm \textit{per se} did not adversely affect the reliability of the model's confidence scores. Deviations from optimal confidence were particularly substantial in 2 conditions (see Table \ref{table:overconfidence}, Figure \ref{fig:compositeBayesfinalconfidence}, and Figure \ref{fig:log_odds_ideal_observer} for visualization in log odds space): firstly, there was marked overconfidence in the Answer Shown - Neutral Advice (OUCS = 0.210).  This necessarily reflects the choice-supportive bias, since no new information was provided. Indeed, we observed significantly greater overconfidence in the Answer Shown condition, compared to the Answer Hidden conditions (all \(ps<0.001\); p values computed by permutation testing (n=10000)). Note that there appears to be a similar level of overconfidence in the Answer Shown - Neutral Advice and the Answer Shown - Same Advice conditions in Figure \ref{fig:compositeBayesfinalconfidence}. However, this is not the case (see Table \ref{table:overconfidence} because the OUCS takes into account the number of trials in each bin -- with the vast majority of trials in the Answer Shown - Same Advice condition falling into the near-maximal range of observed/ideal confidence and therefore being relatively well "calibrated".

There was, however, striking underconfidence in the Answer Hidden - Opposite Advice condition (OUCS = -0.30; see Table \ref{table:overconfidence} and Figure \ref{fig:compositeBayesfinalconfidence}), which reflects the overweighting of opposing advice resulting in a loss of confidence in the initially chosen option. Interestingly, the model was less overconfident in the Answer Shown - Same Advice condition (OUCS = 0.09) - and only marginally overconfident in the Answer Hidden - Same Advice conditions (OUCS = 0.051) -  reflecting more appropriate integration of Same advice into the final observed confidence, compared to Opposite advice.

\begin{table}[h!]
\centering
\renewcommand{\arraystretch}{1.4} % uniform vertical spacing adjustment
\begin{tabular}{||c c c||} 
 \hline
 Advice Type & Display Type & Over/Underconfidence Score \\[0.5ex]
 \hline\hline
\rule{0pt}{3ex}Neutral  & Answer Hidden & 0.001  \\[0.5ex]
 Neutral  & Answer Shown  & 0.210   \\[0.5ex]
 Same    & Answer Hidden & 0.051   \\[0.5ex]
 Same    & Answer Shown  & 0.090   \\[0.5ex]
 Opposite & Answer Hidden & -0.300  \\[0.5ex] 
 Opposite & Answer Shown  & 0.086   \\[0.5ex] 
 \hline
\end{tabular}
\caption{Table showing Over/Underconfidence measures compared to ideal bayesian observer. Positive values indicate overconfidence and negative values indicate underconfidence. Metrics calculated as in \citep{ao2023two}: the OUCS computes the \textit{signed} weighted average difference between the observed final confidence and optimal final confidence across bins of predictions (see Methods for details)}
\label{table:overconfidence}
\end{table}

Thus far, our data provide evidence that in our experimental scenario, LLMs exhibit overconfidence due to a choice-supportive bias and underconfidence due to a marked loss of confidence in the initially chosen option following receipt of contrary evidence. In contrast to the overweighting of opposing information, our data suggest that consistent advice is not significantly overweighted: the model was only marginally overconfident in the Answer Hidden - Same Advice condition, where no choice-supportive bias was at play. 

\begin{figure}[H]
    \centering
    \includegraphics[width=1\linewidth]{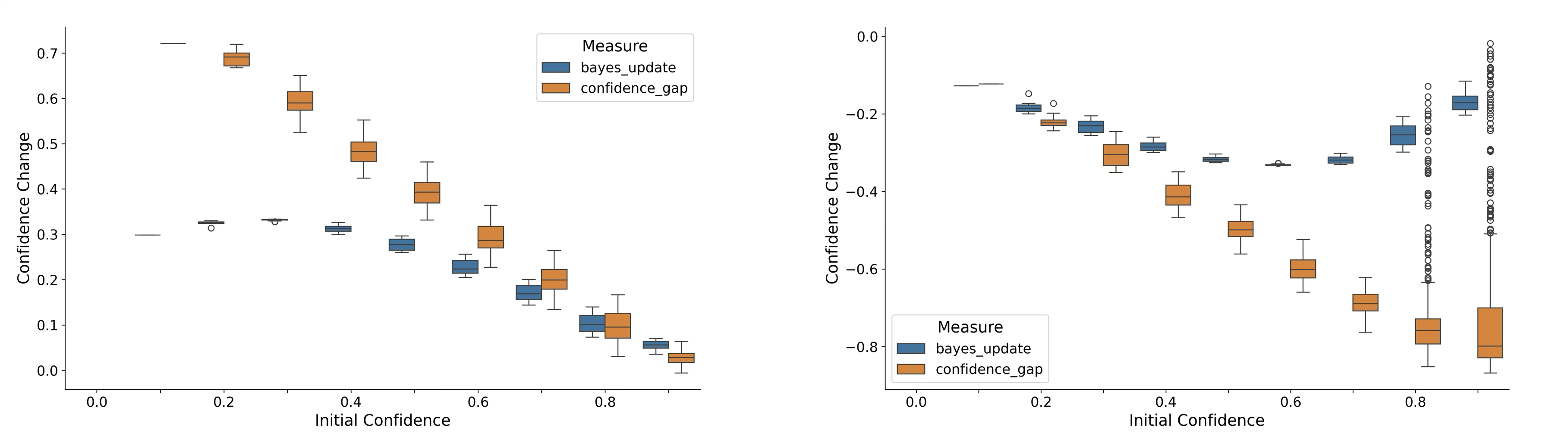}  
    \caption{Comparison of Observed Confidence Updates to Ideal Bayesian Updates in Answer Hidden - Same Advice (left panel) and Answer Hidden - Opposite Advice conditions (right panel). Error bars reflect standard error of the mean. Both plots relate to advice accuracy of 80\%.}
    \label{fig:bayes_update_oppsame80}
\end{figure}

We next sought further evidence that the sensitivity of the LLM to opposing advice was greater than would be expected by an ideal observer. To do this, we compared the observed confidence update that followed the advice given, with that of an ideal bayesian observer. In order to isolate the effect of interest from the influence of choice-supportive bias on confidence, we compared the Answer Hidden - Opposite Advice and Answer Hidden - Same Advice conditions (see Figure \ref{fig:bayes_update_oppsame80}). This revealed that opposing advice was significantly overweighted compared to an ideal observer (observed confidence update / bayesian update ratio = 2.58 (calculated across all accuracies; see Methods), Wilcoxon sign rank test \(p<0.0001\)), and compared to supporting advice (\(p<0.0001\) -- which was weighted only very marginally higher than an ideal observer model (ratio = 1.095, \(p<0.0001\))(see Figures \ref{fig:composite_bayes_update_opp} and \ref{fig:composite_bayes_update_same} for all accuracies). Notably, when we performed a separate analysis on the Answer Shown - Opposite Advice condition -- accounting for the choice-supportive confidence boost exhibited in this condition (see Methods for details) -- this also revealed an overweighting ratio of 2.0 (\(p<0.0001\), Wilcoxon test). 

A priori, one might reason that a model might be unwilling to update its beliefs when information is presented that is contrary to its general knowledge of the world stored in its weights. Whilst in any case, one would expect such conservatism to be expressed in its prior (i.e. initial confidence), our data suggest just the opposite: the model is overly sensitive to contrary information and carries out too large of a confidence update as a consequence. 

Thus far we have compared the \textit{magnitude} of confidence updates to an ideal observer. We next compared the \textit{profile} of observed confidence updates to those of an ideal observer, focusing on \textit{absolute} confidence updates. Notably, a strong prior provides relative protection against contrary advice -- that is the \textit{relative} reduction in confidence (i.e. \% decrease in prior) is highest with high initial priors and linearly decreasing as a function of the prior -- the \textit{absolute} reduction in confidence shows a U-shape (or inverted U-shape) profile (see Figure \ref{fig:bayes_update_oppsame80}). Notably, the profile of the observed \textit{absolute} confidence updates in the Opposite and Same Advice conditions (displayed in Figure \ref{fig:bayes_update_oppsame80}; see Figures \ref{fig:composite_bayes_update_opp} and \ref{fig:composite_bayes_update_same} for plots of all accuracies) deviated qualitatively from the U-shaped profile of Bayesian updates. 

\subsection{Choice-supportive Bias and Overweighting of Contrary Advice in Other Models}
Thus far, we have focussed our analyses on Gemma 3 12B. We next examined the behavior of several other models in the context of the latitude task. We found that Gemma 3 12B, GPT4o and GPTo1 preview also exhibited a choice-supportive bias -- preferentially sticking with their choice when it was visible to them, compared to when it was hidden (see Supplemental Information for details). We also found underconfidence associated with overweighting of opposing advice compared to an ideal Bayesian observer in these models -- as well as deviation from the profile of an ideal observer -- with the exception of GPT o1-preview which we were unable to examine due to the inaccessibility of logits. Whilst DeepSeek 7B performed comparably at the task itself, it did not exhibit the ability to follow advice of the other LLM (unlike the other models tested) -- curtailing further analyses (see Supplemental Information for details). 

\subsection{Choice-Supportive Bias: Additional Experiments}
Our data provide strong evidence that LLMs exhibit a strong choice-supportive bias - tending to reinforce its belief in the correctness of its answer when its answer was visible in the context. Next, we considered 3 possible contributory factors to the observed choice-supportive bias.  One consequence of the choice-supportive bias exhibited by the LLM is a reluctance to change its mind - even when opposing advice is provided. Whilst this could not explain the observed confidence boost produced by the visibility of the initial answer, we considered whether reduced change of mind rate might, in part, be due to the model simply copying its original answer as its final choice -- without even attending to the question tokens present in the second prompt. To address this issue, we conducted an additional experiment, in which there was an extra condition in which the initial answer of the LLM was always replaced by the wrong answer (called the Answer Wrong condition) - as before with varying advice type and advice accuracy. Importantly, this wrong answer was always visible (i.e. as in the Answer Shown condition). Our reasoning was that if the LLM was merely copying its original answer, and ignoring the question tokens, it should exhibit a similar reluctance to change its initial answer in the Answer Wrong condition as observed in the Answer Shown condition reported previously -- despite its initial answer always being wrong in this situation. The LLM's task was simply to judge which of the 2 two-digit numbers in a question was even. We used this simple task so that performance would be at ceiling (i.e. 100\%), guaranteeing that the initial answer in the Answer Wrong condition would always be incorrect.  

\begin{figure}[H]
    \centering
    \includegraphics[width=0.5\linewidth]{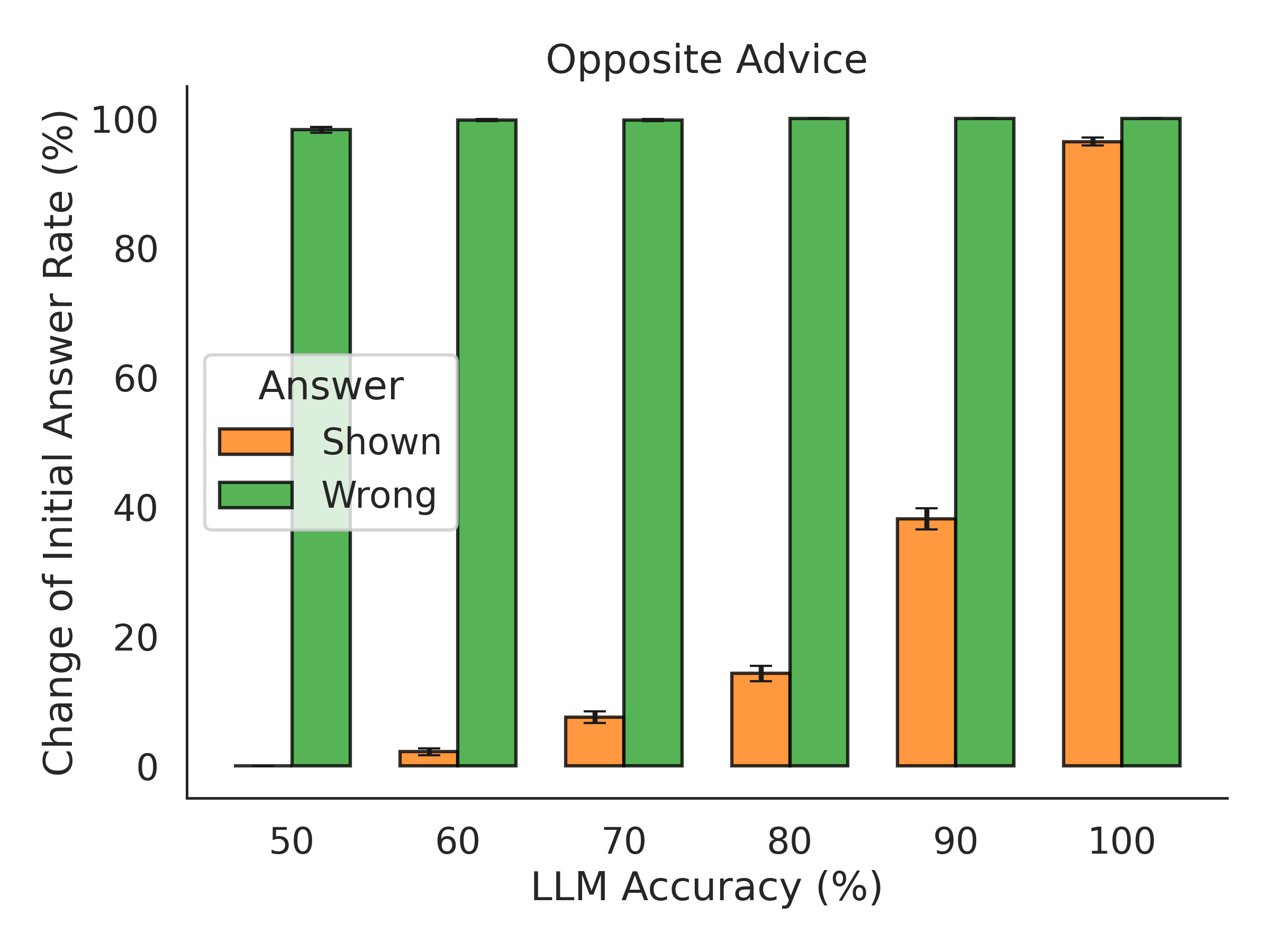}  
    \caption{Additional experiment to investigate if the reduced change of mind rate associated with the choice-supportive bias is driven by copying of the initial answer. The task in this experiment was simply to judge which of two 2-digit numbers was even. In the Answer Wrong condition, the initial answer of the answering LLM was always replaced by the wrong answer. Importantly, this wrong answer was always visible (i.e. as in the Answer Shown condition). As before, in the Answer Shown condition, the initial answer of the answering LLM was revealed in the second prompt. Significantly higher change of initial answer rate in the Answer Wrong condition, as compared to Answer Shown condition argues against possibility of answer copying. Error bars reflect standard error of the mean. Also see Figure \ref{fig:COM_oddevenhog_composite} for all conditions}
    \label{fig:COM_oddevenhog_opp}
\end{figure}

We found a significantly greater change of initial answer rate in the Answer Wrong condition, as compared to the Shown condition (mean change of initial answer rate (collapsed across advice type): 74.5\% and 8.8\%, respectively; \(p<0.0001\)(see Figures \ref{fig:COM_oddevenhog_opp} and \ref{fig:COM_oddevenhog_composite}). By demonstrating that the change of initial answer rate is modulated by the correctness of the initial answer (0\% and 100\% in Answer Wrong and Shown conditions, respectively, this provides compelling evidence that at the time of the second prompt, the LLM does indeed process the question tokens alongside its original answer (when visible). As such, the findings from this experiment argue strongly against the notion that the reduced change of mind rate observed in the main experiment is (i.e. the choice-supportive bias) is simply due to the answering LLM copying its original answer. 

We also considered another factor that could contribute to the observed choice-supportive bias: that in-context information (i.e. the original answer of the LLM when visible) dominates over information that must be retrieved from weights (i.e. a process necessary to answer the question from scratch). To assess this possibility, we conducted an experiment where the LLM had to answer a binary question about the latitude of cities - as in the main latitude experiment - but with a critical difference: the cities (denoted by 4 letter arbitrary strings) were fictitious, and all the necessary information to answer the questions was present in-context (i.e. 400 city-latitude pairs; performance was well matched to that in the main latitude dataset at 76.0\%). Critically, we found a substantial choice-supportive bias in this experiment (no significant difference in magnitude of bias from original experiment: \(p>0.1\)). Specifically, there was significantly less tendency to change mind in the Answer Shown condition (mean change of mind rate = 17.2\%, collapsed across advice type), as compared to the Answer Hidden condition (mean change of mind rate = 33.2\%; significant difference between Shown and Hidden \(p<0.0001\); see Figure \ref{fig:COM_icx_composite}). Given that all information was in-context in this experiment -- specifically, the original answer of the LLM (when visible) as well as the information necessary to answer the question -- this finding argues against the notion that the choice-supportive bias is due to the dominance of in-context information, over in-weights information.  

\begin{figure}[H]
    \centering
    \includegraphics[width=0.5\linewidth]{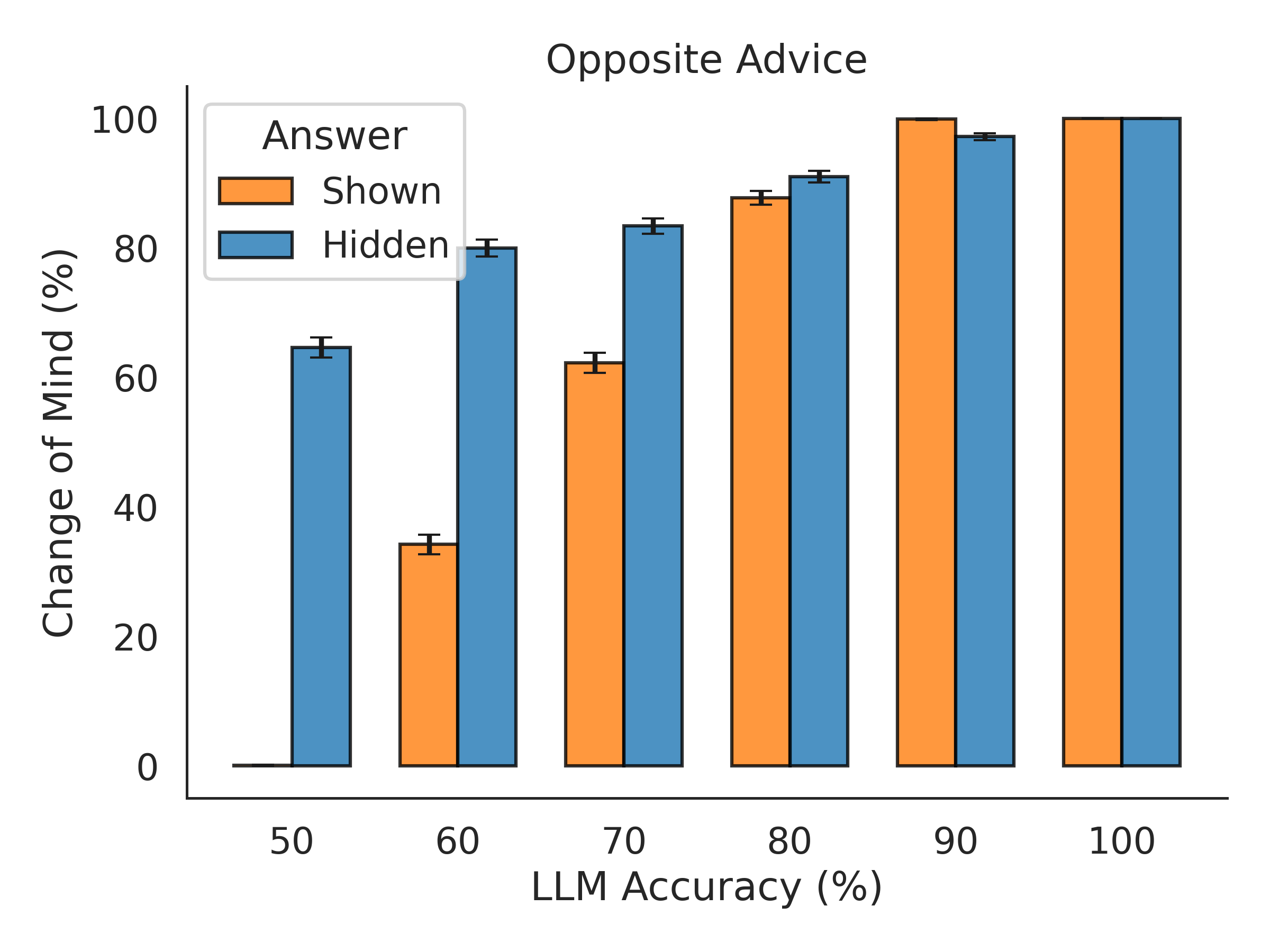}  
    \caption{Results from experiment where the answering LLM was told that the original answer visible within the second prompt came from a different LLM ("of similar size in billions of parameters"; see Methods). The size of the choice-supportive bias was vastly reduced as compared to the original experiment, providing evidence that the bias is driven by the answering LLM sticking to \textit{its own answer}. Error bars reflect standard error of the mean. See text for details}
    \label{fig:COM_LLM1frame_opposite}
\end{figure}

We next asked whether the emergence of a choice-supportive bias is critically dependent on the identity of the agent, whose original answer is visible to the LLM. Specifically, one hypothesis is that a choice-supportive bias arises because the LLM is irrationally tied to justifying and sticking with \textit{its own answer}\citep{henkel2007memory} -- but this effect does not arise if the answer visible to the LLM is known to come from a different actor (i.e. another LLM). To test this hypothesis, we conducted an experiment where we assessed the choice-supportive bias that arose when the answering LLM was told that the original answer came from a different LLM (``of similar size in billions of parameters''; see Supplemental Methods). We found that the tendency of the LLM to change its mind was broadly similar in the Answer Shown and Hidden conditions: mean change of mind rate was equal to 31.3\% and 33.2\% in the Hidden and Shown conditions, respectively (no significant difference, \(p>0.1\); see Figure \ref{fig:COM_LLM1frame_opposite}). As such, no choice-supportive bias was observed in this experiment. This finding provides strong evidence that the LLM shows an irrational bias to stick with the original answer, as long as it is its own -- and not that of another actor. 

\subsection{Modelling Choices and Confidence in a New Dataset: Transfer analysis}
Thus far, we have identified the roots of over/underconfidence in LLMs -- in particular, a choice-supportive bias that causes the model to be overconfident and reluctant to change its mind, and a marked underconfidence in the face of opposing information. Next, we asked whether we could model the behavior of LLMs, specifically Gemma 3 12B, using these simple principles -- that is the influence of the prior on subsequent confidence/change of mind, the overweighting of contrary information, and the visibility of the initial answer -- in a fashion that would allow us to capture their choices and confidence in a novel setting. 

To achieve this, we first modelled choices and confidence in a 4-way multiple choice version of the SimpleQA dataset (\citep{wei2024measuring}; see Methods). Our motivation in opting for this scenario, as compared to the binary choice latitude dataset, was to test the ability of the principles uncovered in the binary scenario to generalize to a slightly more complex choice setting. Further, this 4-choice setting allowed us to assess the transferability of the model beyond the factuality domain (i.e. latitude dataset, SimpleQA) to the reasoning domain (i.e. GSM-MC \citep{zhang2024multiple}; a multiple choice version of the GSM8k dataset).  We modeled the LLM's confidence ratings using a Bayesian beta regression (see Supplemental Methods), which is particularly suitable for bounded outcome variables, such as confidence scores ranging between 0 and 1. 

The model (see Supplemental Methods) had final confidence in the initially chosen option as its dependent measure and included the following predictors: the initial confidence in the initially chosen option (\textit{prior}), a composite measure incorporating the type and accuracy of advice received (\textit{advice received}; see Supplemental Methods for calculation), and a binary indicator for whether the original answer of the LLM was visually displayed or hidden (\textit{shown flag}). 

We set up models of increasing complexity, including more of these predictor terms (see Supplemental Methods). Notably, the full model (i.e. model D) of confidence in the initially chosen option was unique in incorporating individual weights for opposing and supportive advice, for Shown and Hidden conditions separately. This allowed us to capture i) the effect of the initial answer being visible (i.e. choice-supportive bias) ii) the overweighting of opposing advice. For completeness, we also modelled final confidence in the \textit{final chosen option} in a similar way. To model the factors influencing the tendency of the model to change its mind, we used a logistic regression model -- whose predictor variables were similar to the confidence models (see Supplemental Methods). 

We first performed a model comparison considering models of different complexity (see Supplemental Methods), and compared their ability to fit data across both continuous (i.e. confidence) and binary (i.e. change of mind (COM) data. We found robust support for model D both in terms of expected log predictive density leave-one-out(ELPD-LOO; see Supplemental Methods), and its ability to fit a held out set of 10k questions from the SimpleQA dataset (see Figure \ref{fig:heldout_modelfit_QA}). Model D demonstrated the highest predictive accuracy (ELPD-LOO = 17036, SE = 226), significantly outperforming Model C (ELPD-LOO = 14614, SE = 198), with a difference of 2422 points (see Figure \ref{fig:modelcomparison}). Given the magnitude of this difference compared to the standard errors, Model D is considered credibly superior to Model C in predictive terms.

Across all 3 models (i.e. for confidence in the initial option, ultimate option, change of mind data), we found a significant effect of the shown flag -- that is a choice-supportive bias (shared parameter across models: $M = 1.35$ $95\%\ \text{HDI} = [1.33, 1.37]$). Further we found a significant effect of the prior (shared parameter across models: $M = 1.138$ $95\%\ \text{HDI} = [1.09, 1.18]$): confirming the link between this variable and final confidence and tendency to change mind. The posterior analysis also revealed a significant effect of contrary advice in both the Answer Hidden - Opposite Advice ($\text{Mean } = 3.13$, $95\%\ \text{HDI} = [3.04, 3.22]$) and Answer Shown - Opposite Advice ($\text{Mean} = 2.82$, $95\%\ \text{HDI} = [2.76, 2.89]$) conditions (see Table \ref{tab:params_initial}). 

\begin{figure}[H]
    \centering
    \includegraphics[width=1\linewidth]{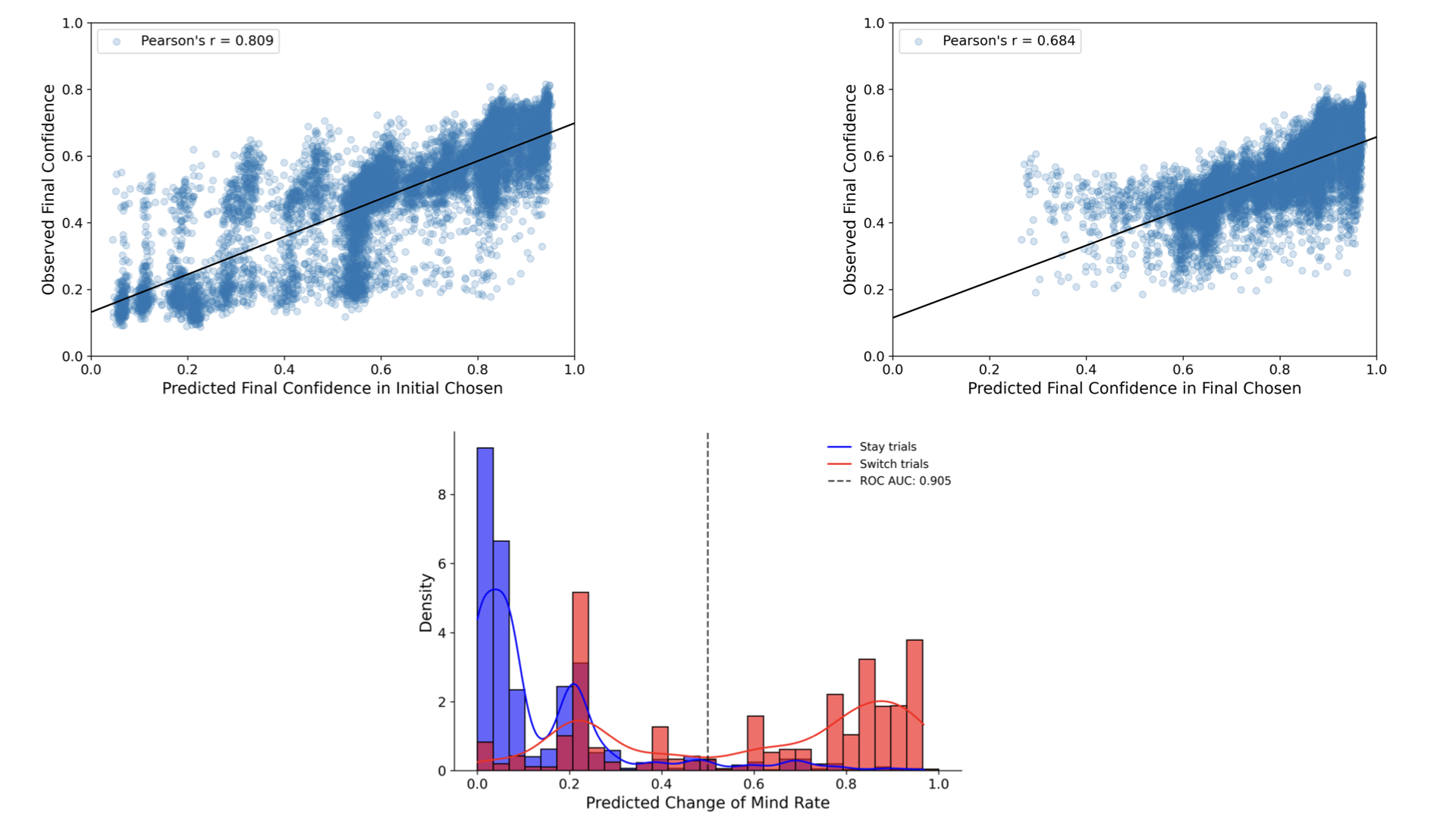}  
    \caption{Model fit on new dataset: multiple choice version of GSM8k dataset (GSM-MC: \citep{zhang2024multiple}). Top left panel: confidence in initially chosen option. Top right panel: confidence in ultimately chosen option. Bottom panel: change of mind data. Each point represents an individual trial. AUROC of change of mind data shown in legend.}
    \label{fig:compositeGSM}
\end{figure}

\begin{figure}[htbp]
    \centering
    \includegraphics[height=7.5cm]{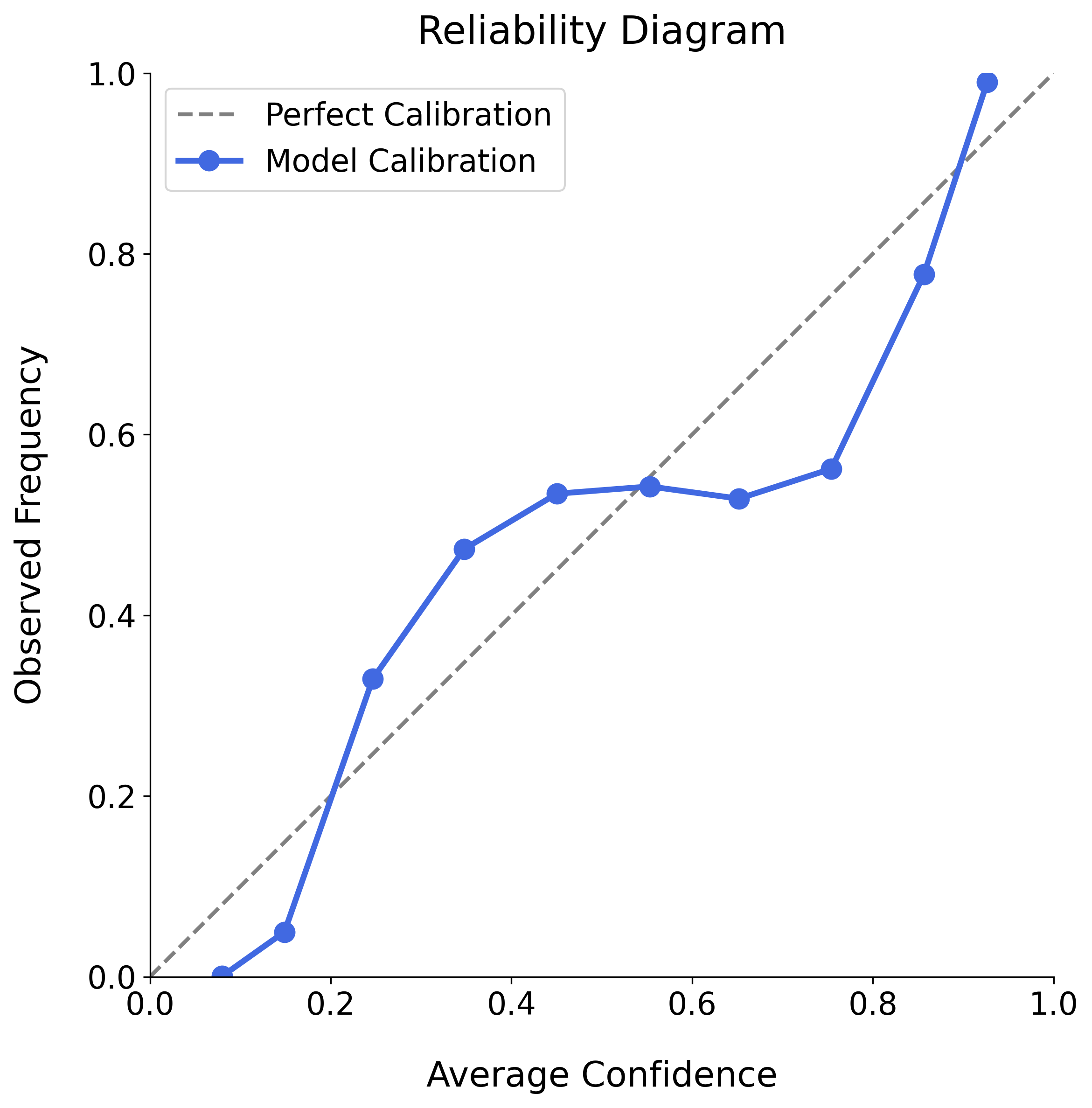}     % 5 centimeters high
    \caption{Reliability diagram: optimal temperature 3.3, resulting in ECE of 0.09, Brier score of 0.15, and AUROC of 0.88. This was obtained using 40k questions which were excluded from the main latitude dataset. Performance of the model on this dataset was 75.6\%}
    \label{fig:reliability}
\end{figure}

\begin{figure}[htbp]
    \centering
    \includegraphics[width=1\linewidth]{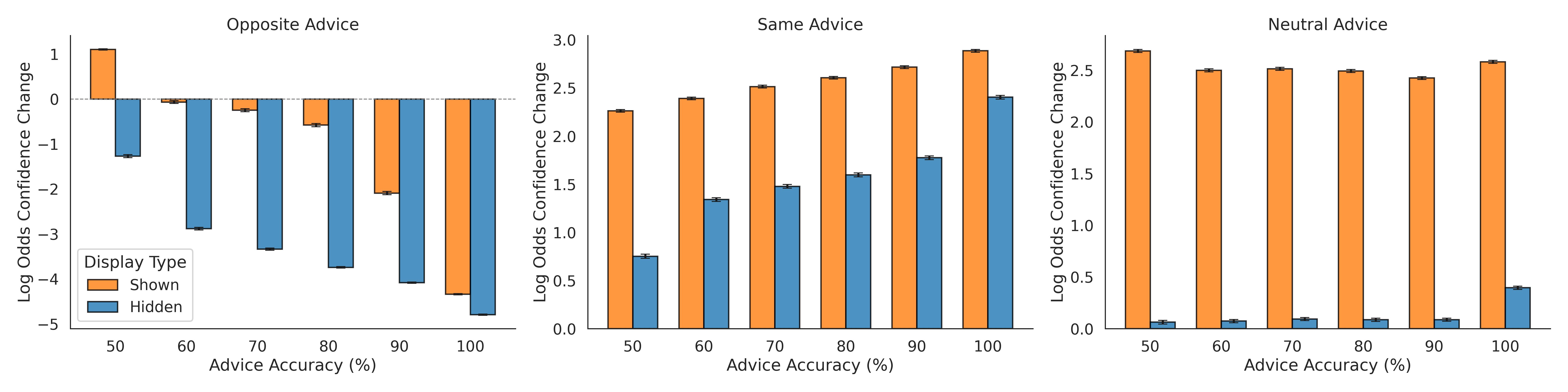}     % 5 centimeters high
    \caption{Confidence Change in Log Odds Space: Effects of Advice type, Advice LLM accuracy. Confidence change refers to the difference between the final confidence in the initially chosen option \textit{regardless of whether it was the ultimately chosen option} and the initial confidence in that option. Error bars reflect standard error of the mean. Note that seemingly large log odds changes (e.g. +3) can represent relatively moderate probability updates when starting from moderate to high confidence (e.g., 75\% → 98\% yields +2.8 log odds change)}
    \label{fig:log_odds_confidence_change}
\end{figure}

\begin{figure}[htbp]
    \centering
    \includegraphics[width=1\linewidth]{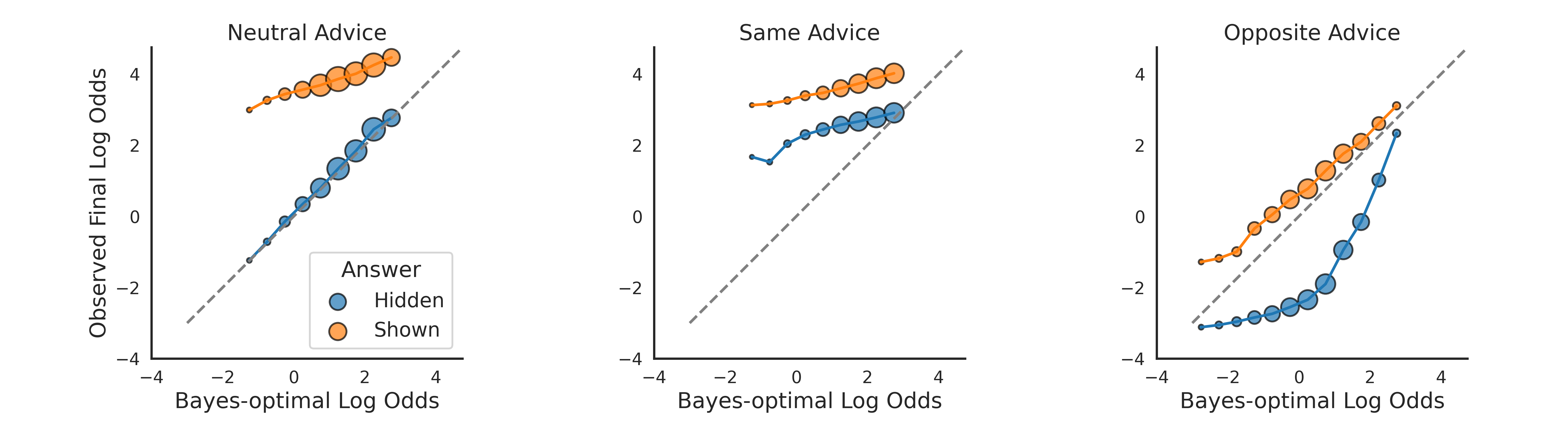}     % 5 centimeters high
    \caption{Relationship between final confidence in the initially chosen option and ideal observer confidence score in log odds space. Note that log odds of 0, 1, 3, 5 correspond to probabilities of 0.5, 0.73 and 0.95, 0.99 respectively. Log odds of -1, -3. -5 correspond to probabilities of 0.27 and 0.05, 0.01 respectively. Data were binned based on the Bayesian log odds values (bin width = 0.5 units), and the average observed log odds confidence was computed for all data points falling within each bin. Note that marker size reflects number of trials. Points above the diagonal line indicate overconfidence whereas below indicate underconfidence, relative to an ideal Bayesian observer. Note that to constrain the data to fit within a reasonable axes scale, 12\% of the data are not shown since they are at the extremes}
    \label{fig:log_odds_ideal_observer}
\end{figure}

\begin{figure}[H]
    \centering
    \includegraphics[width=1\linewidth]{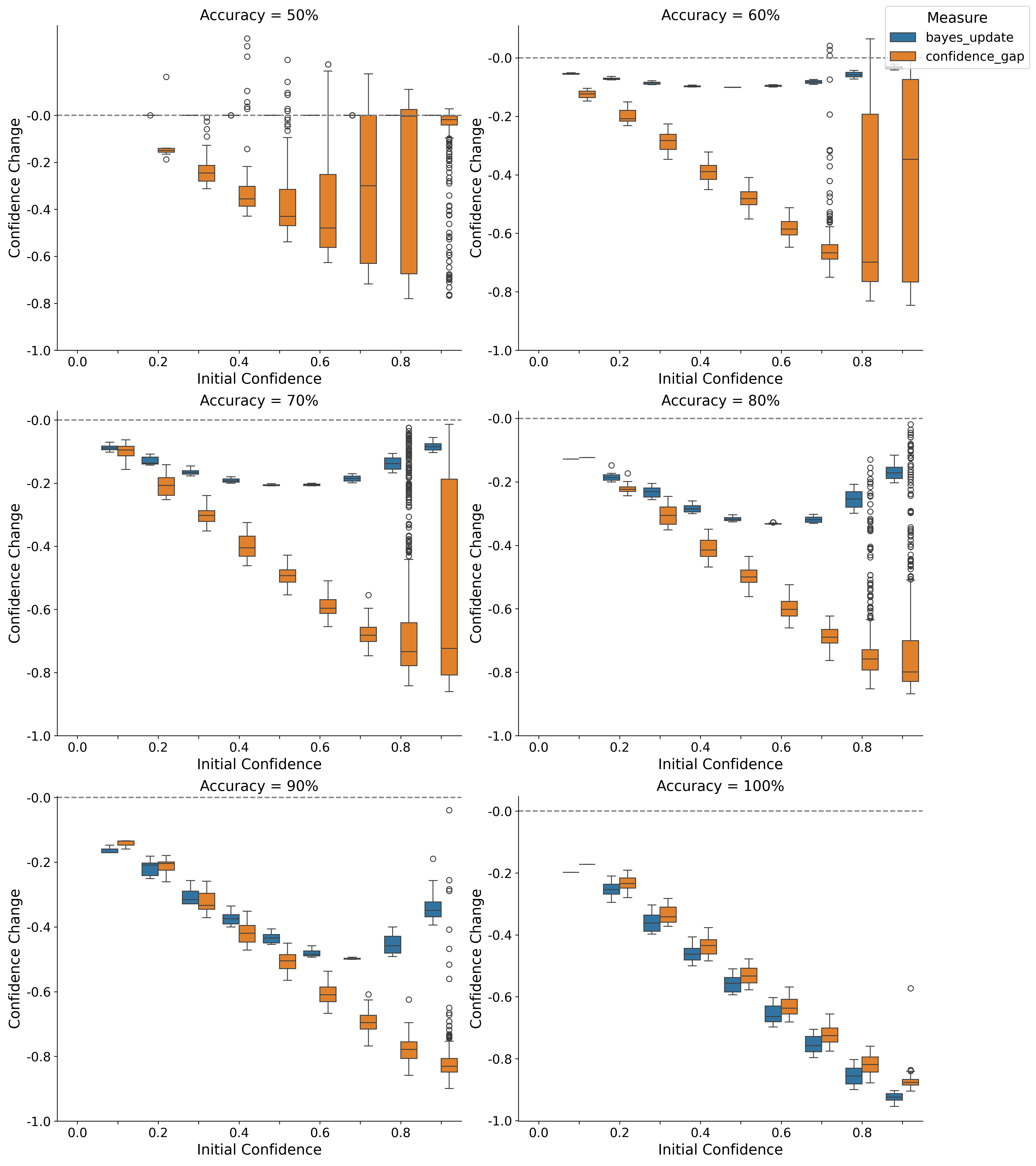}  
    \caption{Comparison of Observed Confidence Updates to Ideal Bayesian Updates in Answer Hidden - Opposite Advice condition shown for all advice accuracies. Note the inverted u-shaped profile of bayesian updates in all conditions except the 50\% and 100\% accuracy conditions. See main text for details.}
    \label{fig:composite_bayes_update_opp}
\end{figure}

\begin{figure}[H]
    \centering
    \includegraphics[width=1\linewidth]{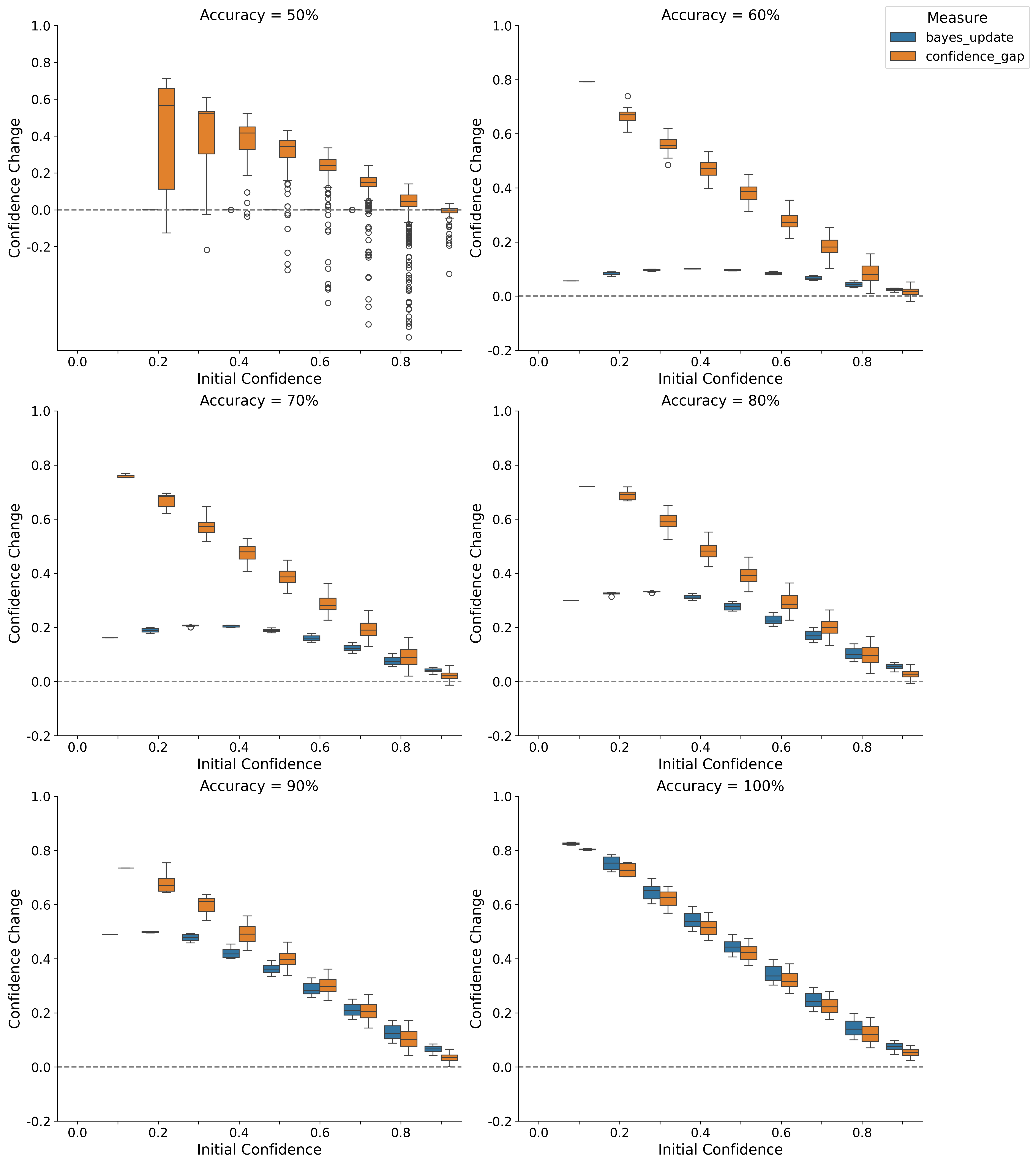}  
    \caption{Comparison of Observed Confidence Updates to Ideal Bayesian Updates in Answer Hidden - Same Advice condition shown for all advice accuracies. Note the inverted u-shaped profile of bayesian updates in all conditions except the 50\% and 100\% accuracy conditions. See main text for details.}
    \label{fig:composite_bayes_update_same}
\end{figure}

Of note, whilst we targeted our modelling to capturing confidence in the initial chosen option, we also modelled change of mind rate and confidence in the ultimately chosen option for completeness (see Figure \ref{fig:compositeGSM}, Supplemental Tables \ref{tab:params_final}, \ref{tab:params_COM}). Across all these measures our full model captured substantial variance in the SimpleQA dataset (see Figure \ref{fig:heldout_modelfit_QA}). We next asked whether the trained model could transfer to a different domain: namely reasoning in the multiple choice version of the GSM8k maths dataset \citep{zhang2024multiple} (cf factuality in the latitude and SimpleQA datasets). Despite the change of domain, we observed that our trained model provided a close fit to the GSM-MC data demonstrating its ability to transfer to a dataset with different task demands (see Figure \ref{fig:compositeGSM}; Model D: ELPD-LOO = 17847, SE 166). 

\section{Discussion}

Our results demonstrate that LLMs deviate from normative behavior in several significant ways: firstly, they exhibit a striking choice-supportive bias which boosts their confidence in their answer, and causes them to stick to it, even in the presence of evidence to the contrary.  Secondly, we show that whilst LLMs do integrate new information into their beliefs, they do so in a fashion that is not optimal: they show a profile of confidence updates that deviates from an ideal observer, and markedly overweight opposing advice resulting in marked loss of confidence in their initial answer. We further demonstrate that a parsimonious model of confidence and change of mind rate developed using a factuality dataset is able to transfer and model LLM beliefs and behavior in a dataset typically used to assess reasoning.

We identify that when LLMs can view their initial answer, this results in a boost of confidence in the correctness of their choice despite the absence of any relevant new information. This choice-supportive bias causes them to have a propensity to stick with their initial choice, reducing the flexibility to change their mind when contrary advice is provided. A unique aspect of our experimental design was instrumental in revealing this choice-supportive bias: we obtained confidence estimates both at the time of initial choice and at the time of final choice, following advice. Critically, final confidence estimates were obtained not only when the answering LLM could view its initial answer (i.e. the Answer Shown condition), but also where they were not influenced by knowledge/memory of the initial answer (i.e. the Answer Hidden condition) -- a experimental manipulation that would not be possible in human participants. 

We asked what factors might drive the observed choice-supportive bias. Firstly, we examined and excluded the possibility that the LLM's reluctance to change its mind when its initial answer was visible could in part be due to simple copying. Second, we evaluated whether the choice-supportive bias might arise due to a dominance of in-context information (i.e. the initial answer when visible) over in-weights information (i.e. knowledge of the city latitude)-- which could engender a greater reliance on the former. To assess this, we conducted an experiment involving fictitious city-latitude pairs where all relevant information necessary to answer the question was presented in-context. The results obtained in this setting argue strongly against the notion that a choice-supportive bias arises due to the preferential treatment of in-context information. Finally, we tested whether the choice-supportive bias required the initial answer to originate from the agent itself. In an experiment where the initial answer was attributed to another agent (i.e. another LLM), we found that the effect was abolished under these conditions. This mirror findings in humans \citep{henkel2007memory} and provides support for the hypothesis that a choice-supportive bias arises specifically from a drive for self-consistency \citep{stocker2007bayesian} -- rather than mere anchoring to any prior response .

This choice-supportive bias echoes similar findings in the cognitive science literature \citep{stocker2007bayesian, jazayeri2007new, sharot2009choice, henkel2007memory}. In one study \citep{jazayeri2007new}, participants were first asked to make a binary choice about the direction of coherence of a field of moving dots. Subsequently, they were asked to estimate the actual direction of the moving dots. They found that participants' direction estimates were heavily biased by the initial categorical decision -- reflecting a choice-supportive bias. Notably, these observations were well accounted for by a model which commits to a choice once made, eliminating the need to maintain a full probability distribution over all hypotheses \citep{stocker2007bayesian}. Although this process sacrifices optimality for simplicity, it serves to conserve mental resources and ensures self consistency -- viewed to be an imperative given that actions in the world usually can't be undone, and future planning is rendered less complex. In other contexts, such a choice-supportive bias may serve to minimize regret \citep{sharot2009choice, henkel2007memory}. As such, participants tend to irrationally strengthen their conviction in the chosen option, and devalue the foregone option -- resulting in a wider gap between 2 initially desirable options \citep{sharot2009choice}. 

Notably, we found evidence of a choice-supportive bias not only on behavior (i.e. change of mind), but also subsequent confidence estimates. In the metacognition literature, previous work has documented the influence of choices on subsequent confidence estimates, potentially mediated by links between the motor system and prefrontal cortical areas mediating confidence formation \citep{siedlecka2016but, wokke2019action, fleming2015action, rollwage2021confirmation}. A theoretical perspective on these findings is furnished by ``second-order'' accounts of metacognition, in which one's own choices are used as proxies for the underlying evidence when inferring whether a decision is likely to be correct \citep{fleming2017self}. These parallels between human cognition and LLM behavior suggest that choice-supportive biases reflect a fundamental computational strategy -- maintaining self-consistency at the expense of optimal updating -- that emerges in both biological and artificial systems.

Closely related to the notion of a choice-supportive bias is the phenomenon of a confirmation bias, where information that is consistent with one's choice is afforded priviledged status over information that is contrary \citep{rollwage2021confirmation}. Indeed, humans have been shown to actively sample information that is consistent with prior choices and beliefs \citep{kaanders2022humans}. However, our study provides evidence that LLMs do not show a confirmation bias, at least under these experimental conditions. Specifically, we found that overconfidence was explained by a choice-supportive bias -- and \textit{not} by overweighting of consistent advice. Whilst our failure to find overweighting of consistent information in the Same Advice Answer Shown condition can be explained by a ceiling effect (i.e. confidence scores were greater than 0.9 for most of the trials in this condition; see Figure \ref{fig:compositeBayesfinalconfidence}) -- this was not the case for the Same Advice Answer Hidden condition, where no overweighting was observed. Whether LLMs show a confirmation bias under different experimental conditions remains an open question. 

Contrary to a confirmation bias, we found that LLMs overweight opposing rather than supportive advice, both when the initial answer of the model was visible and hidden from the model.  One explanation for the oversensitivity to opposing advice we observe is that the training of models with reinforcement learning with human feedback \citep{christiano2017deep, ouyang2022training} -- a technique that aims to align a model's answers with human preferences -- may encourage models to be overly deferential to the preferences or advice of another user or agent, a phenomenon known as sycophancy \citep{sicilia2024accounting, perez2023discovering, sharma2023towards}. Indeed, prior work has shown that RLHF can increase sycophantic behavior \citep{perez2023discovering}, suggesting it could be interesting for future work to examine whether models trained without RLHF exhibit different patterns of confidence updating in our paradigm. Further, whilst a previous study \citep{sicilia2024accounting} showed that LLMs exhibited a sycophantic tendency -- with their accuracy and calibration being influenced by user suggestions --- they did not measure over- or underconfidence, as we do by comparing initial and final confidence in the model's original answer.  

Our results, therefore, reveal a more nuanced pattern than simple sycophancy. Sycophancy is typically characterized by a symmetrical overweighting of both agreeing and dissenting user input. In contrast, we find strikingly asymmetric sensitivity to opposing and supportive advice -- with overweighting of the former but not the latter. Further, we show that the tendency to change mind is not overarching in the face of opposing advice, but modulated by the confidence associated with the LLM's answer. Lastly, we demonstrate that the visibility of the LLM's own answer inflates confidence and induces a marked reluctance to change mind -- a finding that is incompatible with a pure sycophancy-based account. Our results, therefore, are consistent with the operation of two distinct mechanisms: a general hypersensitivity to contradictory information that operates regardless of answer visibility, and a choice-supportive bias that arises due to visible prior answers serving as proxies for the underlying evidence that generated them. It will be interesting to examine if our account extends to more naturalistic user-LLM interactions where free form answers are involved.

\bibliography{references}  % This points to references
\section*{Author Contributions}  % or \section*{Acknowledgements} for British spelling
DK conceived the project with input from SMF, and carried out the experiments. LM, JH, MM provided technical support. VP, PV, SMF, BDM, SO, RP advised on the project. DK wrote the paper, with input from SMF, PV, BDM, and VP.

\section*{Acknowledgments}  % or \section*{Acknowledgements} for British spelling
We thank Nathaniel Daw and Stephanie Chan for comments on an earlier version of the paper
\section*{Funding}  % or \section*{Acknowledgements} for British spelling
This was provided by Google DeepMind
\section*{Competing Interests}  % or \section*{Acknowledgements} for British spelling
There are no competing interests
\clearpage  % Flushes all floats and starts new page
\section*{Supplemental Materials}
\section{Supplemental Methods, Supplemental Results, Extended Data Figures and Tables}

\subsection{Supplemental Methods}
\subsubsection{Modelling: Bayesian regression}

We modeled the LLM's confidence ratings using a Bayesian beta regression (in PyMC), which is particularly suitable for bounded outcome variables, such as confidence scores ranging between 0 and 1. We performed this analysis in the context of a 4-way multiple choice version of the SimpleQA dataset (see Methods for details). Our motivation in opting for this scenario, as compared to the binary choice latitude dataset, was i) to test the ability of the principles uncovered in the binary scenario to generalize to a slightly more complex choice setting ii) to allow us to assess the generalizability of the model beyond the factuality domain (i.e. latitude dataset, SimpleQA) to the reasoning domain (i.e. GSM-MC). 

The full model (see model specification below) included predictors for participants' initial confidence in their initially chosen option (\textit{prior}), a composite measure incorporating the type and accuracy of advice given (\textit{advice received}), and a binary indicator for whether the original answer of the LLM was visually displayed or hidden (\textit{shown flag}).

Advice accuracy was calculated by rescaling according to the advice direction to quantify the strength of advice relative to a neutral baseline. This was done to ensure that in the 4-choice scenario where random guessing (i.e. uninformative advice) corresponds to 25\% performance, actual advice accuracy spanned the range from 0-1. Specifically:

\begin{equation}
\text{actual\_advice\_accuracy} = 
\begin{cases}
\dfrac{p_{\text{advice}} - 0.25}{0.75}, & \text{if advice is supportive (advice\_direction = +1)} \\[10pt]
\dfrac{0.25 - p_{\text{advice}}}{0.25}, & \text{if advice is opposing (advice\_direction = -1)} \\[10pt]
0, & \text{if advice is neutral (advice\_direction = 0)}
\end{cases}
\end{equation}

Here, \emph{supportive advice} indicates advice aligned with the answering LLM's initial choice, \emph{opposing advice} suggests an alternative option, and \emph{neutral advice} provides no directional information. After rescaling, actual advice accuracy ranges from 0 (no influence) to 1 (maximum influence).

The linear combination of these predictors was transformed using a sigmoid function to constrain predictions between 0 and 1. A dispersion parameter was introduced to account for variability in confidence ratings across trials. Finally, participants' observed confidence ratings were modeled using a beta likelihood function. The final model thus captures how initial beliefs (i.e. about the probability that an option is correct), external advice, and visual presentation of advice jointly influence participants' final confidence judgments within a bounded response scale. Importantly, note that we also modelled the final confidence in the LLM's ultimate choice (i.e. on the second turn) in an equivalent fashion.

\begin{align}
    L_{conf} &= \beta_0 + \beta_1 \cdot prior + \beta_2 \cdot {actual\_advice_{accuracy}} \cdot advice_{direction} + \beta_3 \cdot shown_{flag} \\
    \beta_n &\sim \mathcal{N}(0,1) \\
    \mu_{conf} &= \sigma(L_{conf}) \\
    \phi &\sim Exponential(1.0) \\
    \alpha_i &= \mu_i \cdot \phi \\
    \beta_i &= (1 - \mu_i) \cdot \phi \\
    confidence_i &\sim Beta(\alpha_i, \beta_i) \\[8pt]
\end{align}

To model the factors influencing the tendency of the model to change its mind, we used a logistic regression model. The full model(see model specification below) estimates the probability that the LLM changes its initial decision based on three predictors: their prior confidence, the advice received, and whether the intial answer was explicitly shown. The linear predictor $L_{COM}$ is modeled as a combination of these predictors. The prior confidence, advice received, and the explicit presentation of the intial answer (``shown flag'') each contribute linearly to this latent variable. A sigmoid (logistic) function transforms this linear predictor into a probability of switching. The observed switching decisions were then modeled using a Bernoulli likelihood function, where the estimated switch probability directly determines the likelihood of observing the LLM's actual choice to switch or stay.
\begin{align}
    L_{switch} &= \beta_0 + \beta_1 \cdot prior + \beta_2 \cdot {actual\_advice_{accuracy}} \cdot advice_{direction} + \beta_3 \cdot shown_{flag} \\
    \beta_n &\sim \mathcal{N}(0, 1) \\
    p_{switch} &= \sigma(-L_{switch}) \\
    switch_{obs} &\sim Bernoulli(p_{switch})
\end{align}

Model specification: All models contained intercept terms. Model A contained one predictor, that is the prior. Model B contained two predictors, the prior and advice received. Model C contained three predictors: the prior, advice received, and a binary indicator of whether the initial answer of the LLM was displayed or hidden. Model D was similar in set up to model C with the following exception: individual weights were used for each advice type and type of observed data (i.e. final confidence in the initial chosen option, final confidence in the ultimate choice, change of mind data). Further, individual weights were used for the Answer Shown - Opposite Advice and Answer Hidden - Opposite Advice condition in the model of confidence in the initially chosen option only. This was motivated by our finding of striking underconfidence of the LLM in the Answer Hidden - Opposite Advice condition, when compared to an ideal observer. 

The models were compared using Expected Log Predictive Density calculated via Leave-One-Out cross-validation (ELPD-LOO). ELPD-LOO provides a measure of predictive accuracy, estimating how well a model generalizes to new, unseen data, with higher values indicating better predictive performance. We assessed overall model fit across all the 3 observed data types. 

\subsubsection{Prompt of Additional Experiment where another LLM provides the initial answer}
Here the initial answer was said to be provided by another LLM. The key changes to the second prompt are highlighted in italics.\\
User-turn2: You can see above the original answer to the question \textit{which was provided by another LLM (LLM 1) of similar size (in billions of parameters) to you}. The accuracy of LLM 1 is unknown. You are also now told that there is another LLM (\textit{LLM 2}) which has been trained on exactly the same dataset. LLM 2 is known to have {x} \% accuracy on this latitude dataset (where 50\% is random chance). What this means is that LLM 2's answer has a {x} \% chance of being correct. You are also told that LLM 2 answered {y}. You can change the original answer of LLM 1 if you want. Respond with either `My final answer is: a or`My final answer is:b'.

\clearpage  % Flushes all floats and starts new page
\subsection{Supplemental Results}
\subsection{Additional Models: Results}
\subsubsection{Gemma 3 27B}
\begin{figure}[H]
    \centering
    \includegraphics[width=1\linewidth]{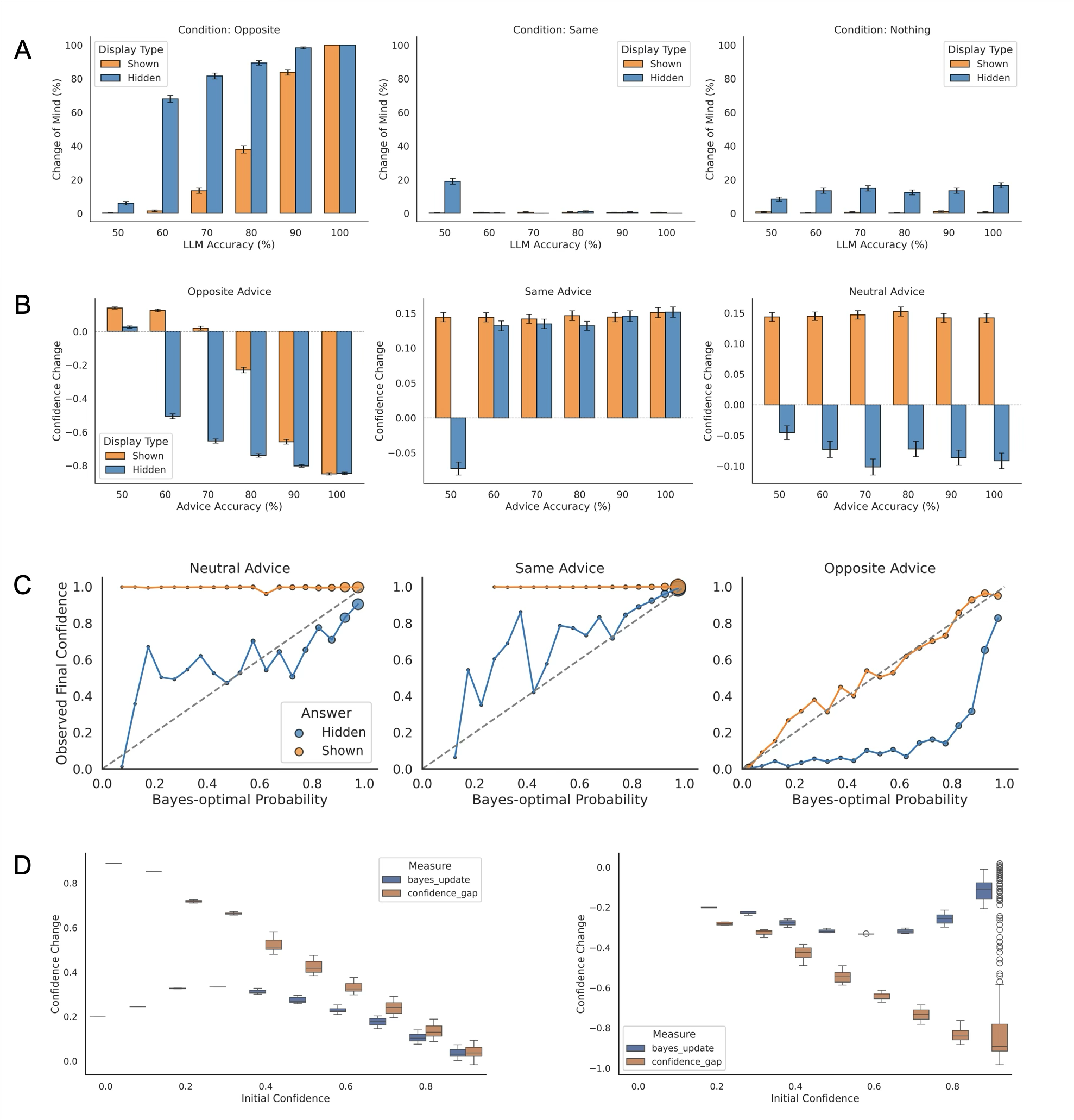}  
    \caption{Results from Gemma 3 27B. A) Effect of Advice Type, Advice Accuracy and Display Type on Change of Mind Rate. Error bars reflect standard error of the mean. B) Effect of Advice Type, Advice Accuracy and Display Type on change in confidence in initially chosen option between first and second turns. Error bars reflect standard error of the mean. C) Relationship between final confidence in the initially chosen option (regardless of whether it was ultimately chosen) and ideal confidence score predicted by a Bayesian observer (see Methods for details) D) Comparison of Observed Confidence Updates to Ideal Bayesian Updates in Answer Hidden - Same Advice (left panel) and Answer Hidden - Opposite Advice conditions (right panel). Error bars reflect standard error of the mean. Both plots relate to advice accuracy of 80\%.}
    \label{fig:Gemma27B_composite_created}
\end{figure}

Initial calibration determined the optimal temperature (2.5; ECE = 0.06), and performance on the standard latitude task was 82.0\%. As for Gemma 3 12B, we found a substantial choice-supportive bias (i.e. 16.7\% difference in change of mind rate averaged across conditions; +0.18 in confidence; see Figure \ref{fig:Gemma27B_composite_created}). We performed the same analyses as for the 12B model. The effect of the Shown condition (i.e. display type) was highly significant in the regression analysis (coeff = -3.4, \(p<0.0001\)). Further, there was a significant negative relationship between the prior and change of mind ($R^2$ = 0.92;\(p<0.0001\)).  

We next examined the advice following ability of the model, first examining tendency to change its mind in relation to the direction of advice and its accuracy. We found a significant effect of advice type: relative to the Neutral Advice baseline condition, there was a reduced tendency to change its mind in the Same Advice condition (coeff = -1.45) and an increased tendency in the Opposite Advice condition (coeff = 2.94; both \(ps<0.0001\))(see Figure \ref{fig:Gemma27B_composite_created}A). We also examined the potential effect of advice accuracy focussing on the Opposite Advice condition, since the majority of change of mind trials occurred in this condition. We found that there was a significant effect of advice accuracy in both the Answer Hidden - Opposite Advice and Answer Shown - Opposite Advice conditions (coeffs = 15.1, 20.4, respectively: both \(ps<0.0001\); See Figure \ref{fig:Gemma27B_composite_created}); though there was a marked difference in tendency to change mind in these conditions (73.8\% vs 39.2\%, respectively).

Our findings on the relationship between display type (Answer Shown or Hidden), advice type and accuracy and confidence in the initially chosen option (i.e. confidence change between first and second turns) paralleled the findings on change of mind (see Figure \ref{fig:Gemma27B_composite_created}B). In a linear regression with confidence change as the dependent variable, we found a significant effect of display type (Shown coeff = 0.20; i.e. implying a +0.20 rise in confidence score when the initial answer was visible \(p<0.0001\)) - revealing the signature of the choice-supportive bias in the confidence ratings. There was also a significant effect of advice type (coeffs = -0.45, and 0.09 for Opposite and Same Advice conditions) and accuracy (\(p<0.0001\)) on confidence. 

Having demonstrated a choice-supportive bias in confidence scores, we next asked whether opposing advice caused larger magnitude confidence decreases compared to the confidence increases produced by supportive advice -- reflecting hypersensitivity to contrary information. We found a significant difference between the magnitude of weights of both the Answer Hidden - Opposite Advice condition (-0.29) and the Answer Shown - Opposite Advice (-0.26) condition versus the Answer Hidden - Same Advice condition (0.12)  on final confidence in the initially chosen option (significant difference, \(p<0.0001\)). We also demonstrated that there was a similar overweighting of opposing advice, as compared the supportive advice, in the Answer Shown - Opposite Advice condition -- providing evidence that opposing advice was overweighted more generally, and not only when the answering LLM's answer was hidden from it at the time of final choice. As for Gemma 3 12B, we observed a significant difference between the weighting of advice in the Answer Hidden - Same Advice and Answer Shown - Same Advice conditions (0.12, 0.05; respectively; difference \(p<0.0001\). 

We next compared the final confidence in the initially chosen option to that of an ideal Bayesian observer. As before, the most marked findings (see Table \ref{table:Gemma27B_overconfidence} and Figure \ref{fig:Gemma27B_composite_created}C) were overconfidence in the Answer Shown - Neutral Advice condition (OUCS = 0.14) -- reflecting the pure choice-supportive bias -- and underconfidence in the Answer Hidden - Opposite Advice condition (OUCS = -0.33), due to loss of confidence in the initially chosen option. 

We also compared the magnitude and profile of confidence updates with that of an ideal observer, following the procedure in the main analyses (see Results). This showed that opposing advice was significantly overweighted compared to an ideal observer (observed confidence update / bayesian update ratio = 3.40 (calculated across all accuracies), Wilcoxon sign rank test \(p<0.0001\)), and compared to supporting advice (\(p<0.0001\) -- which was weighted only very marginally higher than an ideal observer model (ratio = 1.26, \(p<0.0001\)). Further, the profile of the observed confidence updates in the Opposite and Same Advice conditions deviated qualitatively from the profile of bayesian updates (see Figure \ref{fig:Gemma27B_composite_created}D).

\begin{table}[h!]
\centering
\renewcommand{\arraystretch}{1.4} % uniform vertical spacing adjustment
\begin{tabular}{||c c c||} 
 \hline
 Advice Type & Display Type & Over/Underconfidence Score \\[0.5ex]
 \hline\hline
\rule{0pt}{3ex}Neutral  & Answer Hidden & -0.079  \\[0.5ex]
 Neutral  & Answer Shown  & 0.144   \\[0.5ex]
 Same    & Answer Hidden & 0.027   \\[0.5ex]
 Same    & Answer Shown  & 0.069   \\[0.5ex]
 Opposite & Answer Hidden & -0.338  \\[0.5ex] 
 Opposite & Answer Shown  & 0.007   \\[0.5ex] 
 \hline
\end{tabular}
\caption{Table showing Over/Underconfidence measures compared to ideal bayesian observer: Gemma 3 27B. Positive values indicate overconfidence and negative values indicate underconfidence. Metrics calculated as in \citep{ao2023two}}
\label{table:Gemma27B_overconfidence}
\end{table}

\subsubsection{GPT4o}
\begin{figure}[H]
    \centering
    \includegraphics[width=1\linewidth]{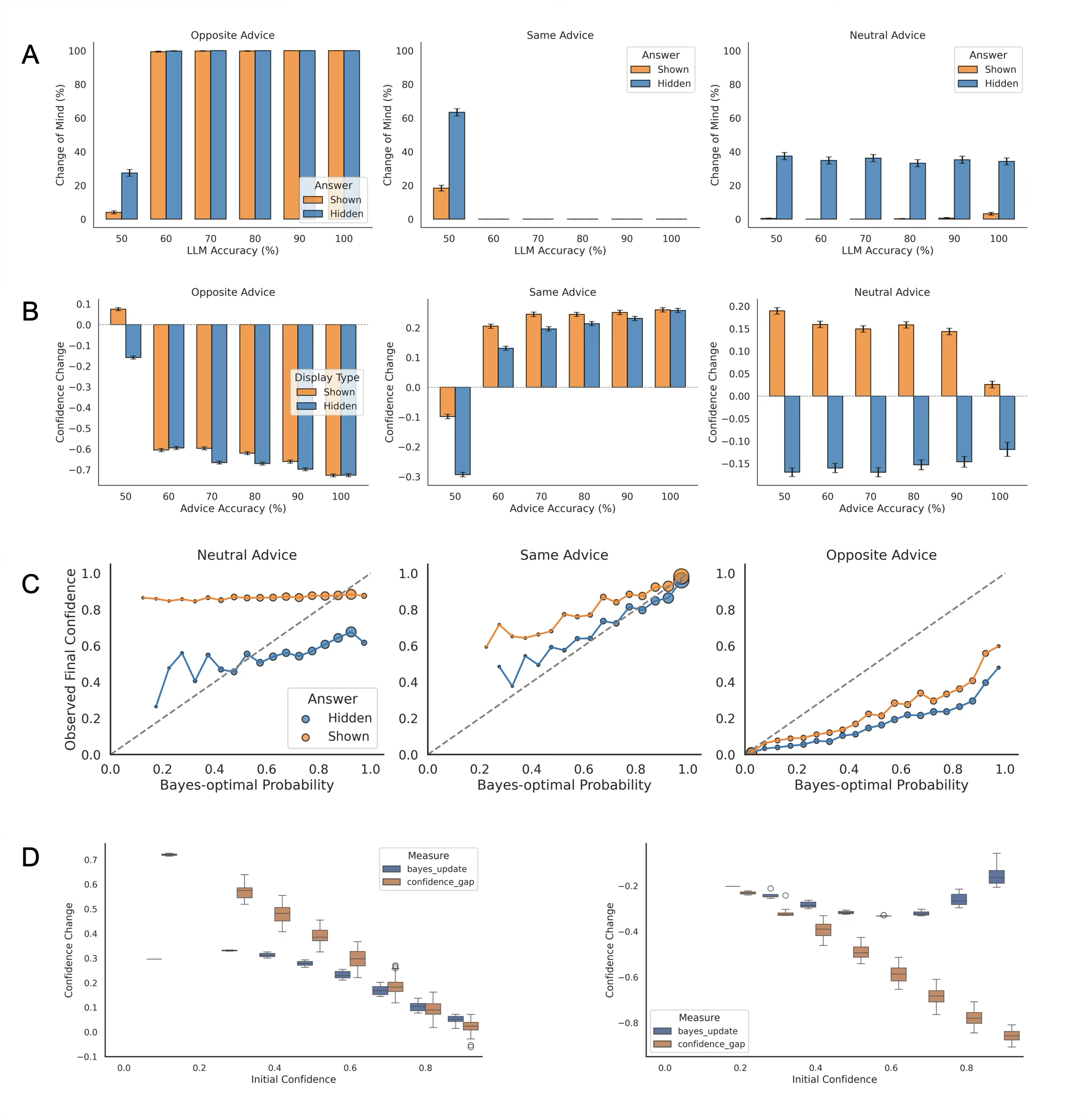}  
    \caption{Results from GPT4o. A) Effect of Advice Type, Advice Accuracy and Display Type on Change of Mind Rate. Error bars reflect standard error of the mean. B) Effect of Advice Type, Advice Accuracy and Display Type on change in confidence in initially chosen option between first and second turns. Error bars reflect standard error of the mean. C) Relationship between final confidence in the initially chosen option (regardless of whether it was ultimately chosen) and ideal confidence score predicted by a bayesian observer (see Methods for details) D) Comparison of Observed Confidence Updates to Ideal Bayesian Updates in Answer Hidden - Same Advice (left panel) and Answer Hidden - Opposite Advice conditions (right panel). Both plots relate to advice accuracy of 80\%.}
    \label{fig:GPT4o_composite_created}
\end{figure}

Calibration showed the optimal temperature was 3.5 (ECE = 0.14). Performance was 93\% on the standard latitude task; hence we tailored the proximity of the foil option to the ground truth option to match performance to Gemma 3 12B (see Methods). Performance on this ``difficult latitude task'' was 77.9\%. There was a significant choice-supportive bias (i.e. 15.3\% difference in change of mind rate averaged across conditions; +0.14 in confidence; see Figure \ref{fig:GPT4o_composite_created}A). We performed the same analyses as for the Gemma models. 

The effect of the Shown condition (i.e. display type) was highly significant in the regression analysis (coeff = -1.32, \(p<0.0001\)). Further, there was a significant negative relationship between the prior and change of mind ($R^2$ = 0.78;\(p<0.0001\)).  

There was a significant effect of advice type: relative to the Neutral Advice baseline condition, there was a reduced tendency to change its mind in the Same Advice condition (coeff = -1.10) and an increased tendency in the Opposite Advice condition (coeff = 3.30; both \(ps<0.0001\))(see Figure \ref{fig:GPT4o_composite_created}). There was a significant effect of advice accuracy in both the Answer Hidden - Opposite Advice and Answer Shown - Opposite Advice conditions (coeffs = 12.1, 11.0, respectively: both \(ps<0.0001\); See Figure \ref{fig:GPT4o_composite_created}). Although there was a marked difference in tendency to change mind in the Neutral Advice conditions (35.2\% vs 0.7\% in Answer Hidden and Answer Shown conditions, respectively), there was a relatively similar change mind rate in the Oppoosite Advice conditions (87.9\% vs 83.8\% in Hidden vs Shown). This was likely due to a ceiling effect due to the heightened sensitivity of the model to the effects of contrary information (see below). 

We found similar effects on confidence as with the Gemma models (see Figure \ref{fig:GPT4o_composite_created}B). A linear regression with confidence change in the initially chosen option as the dependent variable showed a significant effect of display type (Shown coeff = 0.14; i.e. implying a +0.14 rise in confidence score when the initial answer was visible \(p<0.001\)) -  the signature of the choice-supportive bias in the confidence ratings. There was also a significant effect of advice type (coeffs = -0.55, and 0.16 for Opposite and Same Advice conditions) and accuracy (\(p<0.001\)) on confidence. 

We next asked whether opposing advice caused larger magnitude confidence decreases compared to the confidence increases produced by supportive advice -- reflecting hypersensitivity to contrary information. We found a significant difference between the magnitude of weights of both the Answer Hidden - Opposite Advice condition (-0.59) and the Answer Shown - Opposite Advice (-0.66) condition versus the Answer Hidden - Same Advice condition (0.12)  on final confidence in the initially chosen option (\(p<0.0001\)). We also demonstrated that there was a similar overweighting of opposing advice, as compared the supportive advice, in the Answer Shown - Opposite Advice condition -- providing evidence that opposing advice was overweighted more generally, and not only when the answering LLM's answer was hidden from it at the time of final choice. In fact, consistent with the heightened sensitivity to opposing information in GPT4o (see Figure \ref{fig:GPT4o_composite_created} vs left panel of Figure 2)  we found significantly greater overweighting of opposite advice in both Opposite Advice conditions compared to Gemma 3 12B (averaged across both conditions; difference = -0.13, z = -30.5, \(p<0.0001\)). As for the Gemma models, we observed a significant difference between the weighting of advice in the Answer Hidden - Same Advice and Answer Shown - Same Advice conditions (0.12, 0.05; respectively; difference \(p<0.0001\)). 

Comparisons to the ideal Bayesian observer model revealed that the most marked findings (see Figure \ref{fig:GPT4o_composite_created}C and Table \ref{table:GPT4o_overconfidence}) were overconfidence in the Answer Shown - Neutral Advice condition (OUCS = 0.14) -- reflecting the pure choice-supportive bias -- and underconfidence in the Answer Hidden - Opposite Advice condition (OUCS = -0.31). In addition there was underconfidence in the Answer Shown - Opposite Advice condition (OUCS = -0.25) due to the overweighting of opposing information. 

We found that opposing advice was overweighted compared to an ideal observer (observed confidence update / bayesian update ratio = 2.44 (calculated across all accuracies), Wilcoxon sign rank test \(p<0.0001\)), and compared to supporting advice (\(p<0.0001\) -- which was weighted only very marginally higher than an ideal observer model (ratio = 1.08, \(p<0.0001\))(see Figure \ref{fig:GPT4o_composite_created}D). Further, the profile of the observed confidence updates in the Opposite and Same Advice conditions deviated qualitatively from the profile of bayesian updates (see Figure \ref{fig:GPT4o_composite_created}).

\begin{table}[h!]
\centering
\renewcommand{\arraystretch}{1.4} % uniform vertical spacing adjustment
\begin{tabular}{||c c c||} 
 \hline
 Advice Type & Display Type & Over/Underconfidence Score \\[0.5ex]
 \hline\hline
\rule{0pt}{3ex}Neutral  & Answer Hidden & -0.154  \\[0.5ex]
 Neutral  & Answer Shown  & 0.136   \\[0.5ex]
 Same    & Answer Hidden & -0.015   \\[0.5ex]
 Same    & Answer Shown  & 0.047   \\[0.5ex]
 Opposite & Answer Hidden & -0.312  \\[0.5ex] 
 Opposite & Answer Shown  & -0.250   \\[0.5ex] 
 \hline
\end{tabular}
\caption{Table showing Over/Underconfidence measures compared to ideal bayesian observer: GPT4o. Positive values indicate overconfidence and negative values indicate underconfidence. Metrics calculated as in \citep{ao2023two}}
\label{table:GPT4o_overconfidence}
\end{table}

\subsubsection{GPT o1-preview}
We tested GPT o1-preview to specifically assess the presence of a choice-supportive bias in this more powerful model \citep{ai2023gpt}; logits are not accessible through the Open AI API, hence no analysis of confidence was possible. 

\begin{figure}[H]
    \centering
    \includegraphics[width=1\linewidth]{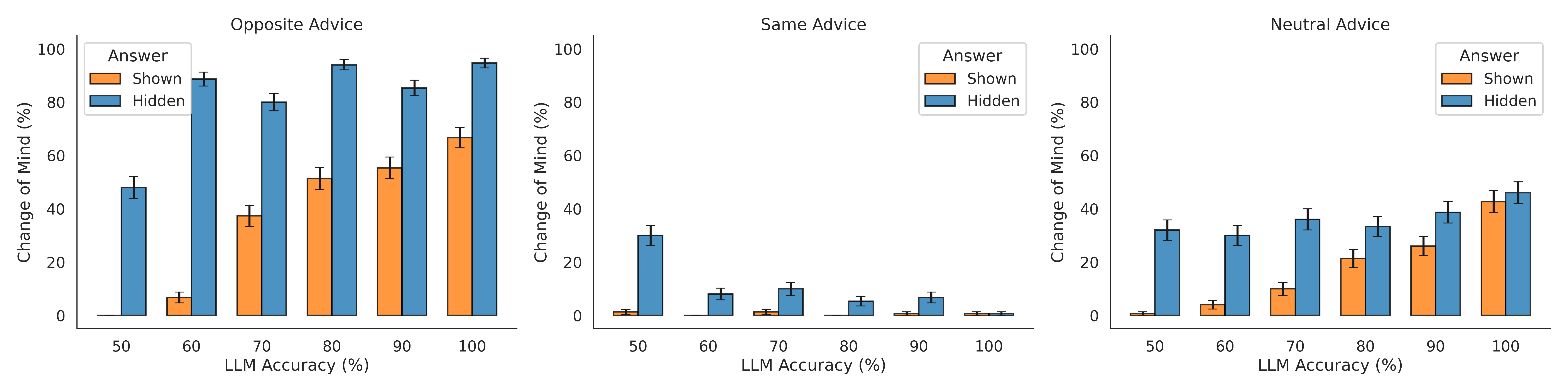}  
    \caption{Results from GPT o1-preview. Effect of Advice Type, Advice Accuracy and Display Type on Change of Mind Rate. Error bars reflect standard error of the mean. }
    \label{fig:GPTo1preview_COM}
\end{figure}

Performance was 78.2\% on the difficult version of the latitude task. The change mind rate was significantly different between the Answer Hidden and Answer Shown conditions (42.6\% vs 18.1\%, respectively; difference = 24.5\% \(p<0.0001\); see Figure \ref{fig:GPTo1preview_COM}). Indeed, the effect of the Shown condition (i.e. display type) was highly significant in the regression analysis (coeff = -1.73, \(p<0.0001\)). These findings provide evidence that this model also shows a choice-supportive bias \citep{henkel2007memory}.

We next tested the efficacy of advice following. We found a significant effect of advice type: relative to the Neutral Advice baseline condition, there was a reduced tendency to change its mind in the Same Advice condition (coeff = -1.86) and an increased tendency in the Opposite Advice condition (coeff = 1.37 ; both \(ps<0.0001\))(see Figure \ref{fig:GPTo1preview_COM}). This finding demonstrates that the answering LLM appropriately integrates the direction of advice to modulate its change of mind rate. 

To examine the potential effect of advice accuracy, we focussed on the Opposite Advice condition, as previously. We found that there was a significant effect of advice accuracy in both the Answer Hidden - Opposite Advice and Answer Shown - Opposite Advice conditions (coeffs = 5.1, 7.5 , respectively: both \(ps<0.0001\); See Figure \ref{fig:GPTo1preview_COM}); though there was a marked difference in tendency to change mind in these conditions (81.8\% vs 36.2\%, respectively).

\subsection{DeepSeek 7B LLM chat}
Finally, we examined the 7B DeepSeek Model. The optimal temperature discovered by calibration was 1.9 (ECE = 0.18). Performance on the standard latitude task was near chance levels; hence we used a division by 8 task (i.e. selecting which of two 5-digit numbers is divisible by 8) since performance was 74.2\% and therefore comparable to the other models on the latitude task.  

\begin{figure}[H]
    \centering
    \includegraphics[width=1\linewidth]{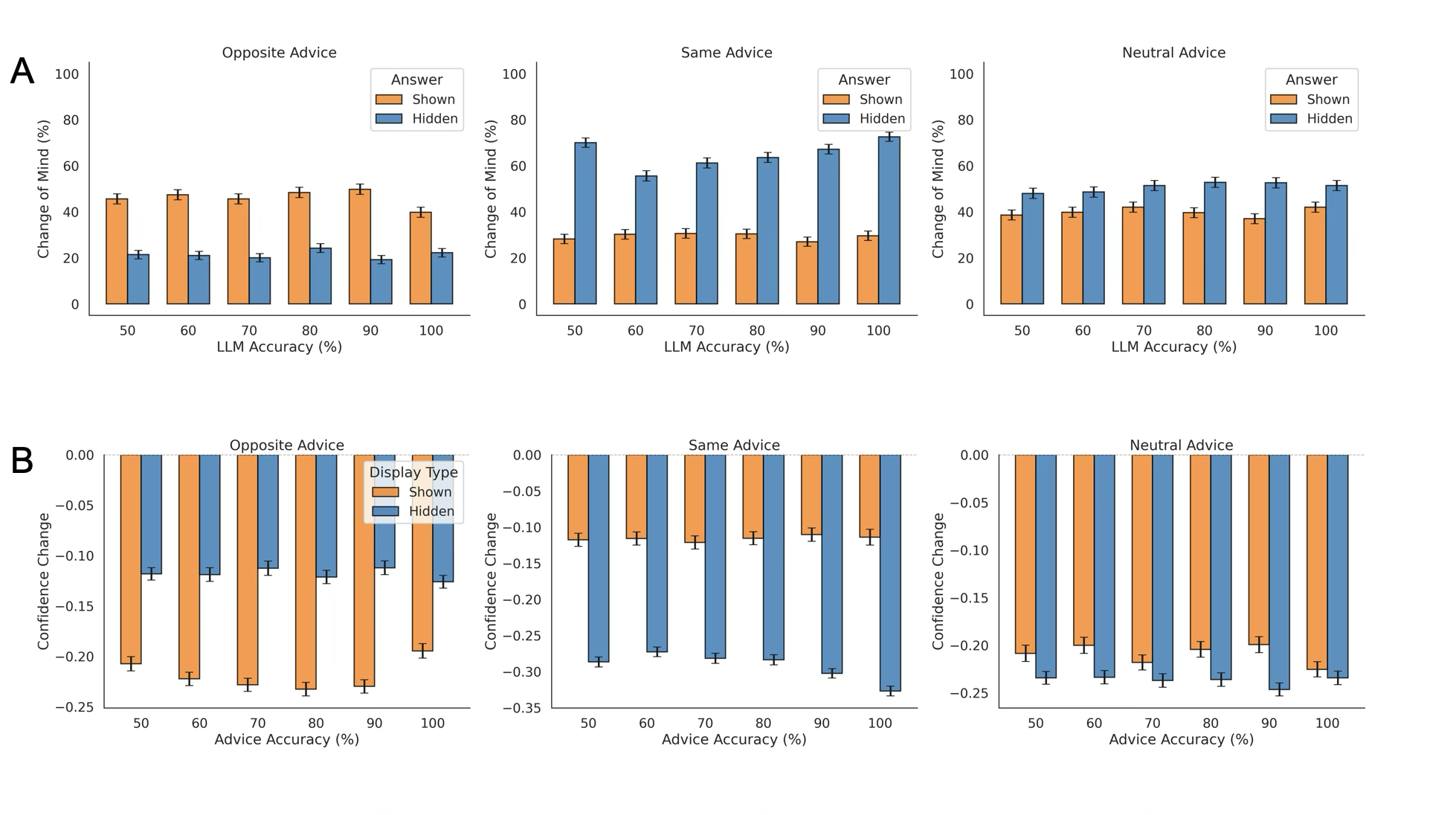}  
    \caption{Results from DeepSeek 7B. Effect of Advice Type, Advice Accuracy and Display Type on Change of Mind Rate. Error bars reflect standard error of the mean}
    \label{fig:DeepSeek_composite_created}
\end{figure}

\begin{figure}[H]
    \centering
    \includegraphics[width=1\linewidth]{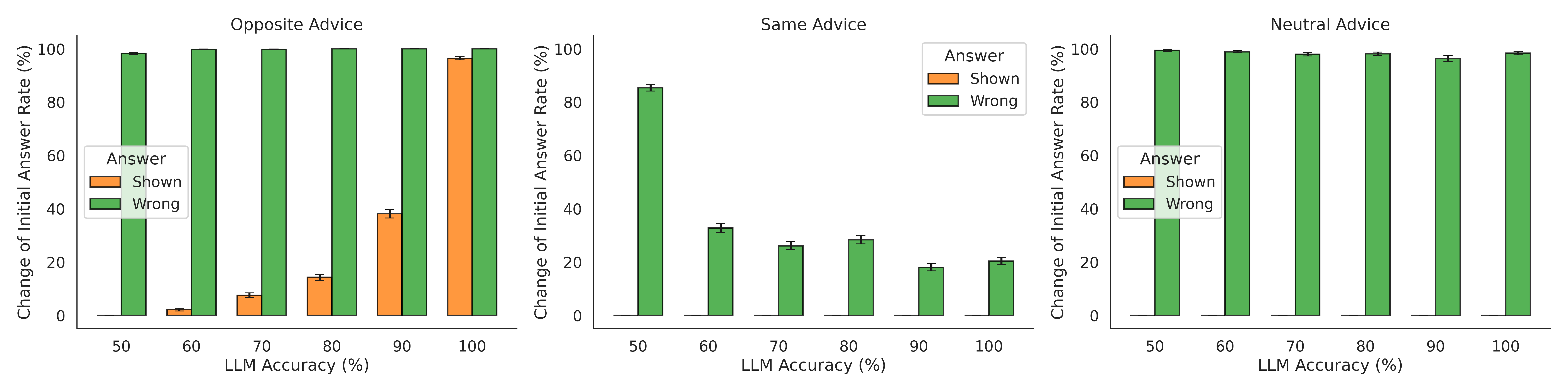}  
    \caption{Additional experiment to investigate if confirmation bias is driven by copying of initial answer. The task in this experiment was simply to judge which of two 2-digit numbers was even. In the Answer Wrong condition, the initial answer of the answering LLM was always replaced by the wrong answer. As before, in the Answer Shown condition, the initial answer of the answering LLM was revealed in the second prompt. Significantly higher change of initial answer rate in the Hog condition, as compared to Shown condition argues against possibility of answer copying. Error bars reflect standard error of the mean.}
    \label{fig:COM_oddevenhog_composite}
\end{figure}

\begin{figure}[H]
    \centering
    \includegraphics[width=1\linewidth]{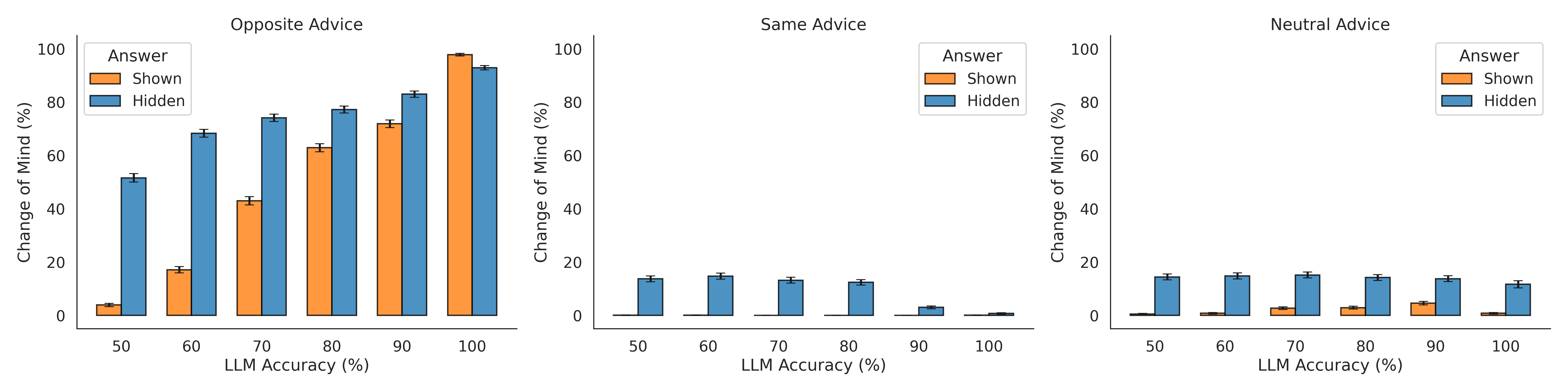}  
    \caption{Results of In-context experiment: in this experiment, all the required knowledge (i.e. city:latitude pairs) were presented in-context. Error bars reflect standard error of the mean. We found a confirmation bias, but one that was significantly lower than in the original experiment (see main text for details)}
    \label{fig:COM_icx_composite}
\end{figure}

\begin{figure}[htbp]
    \centering
    \includegraphics[width=0.5\linewidth]{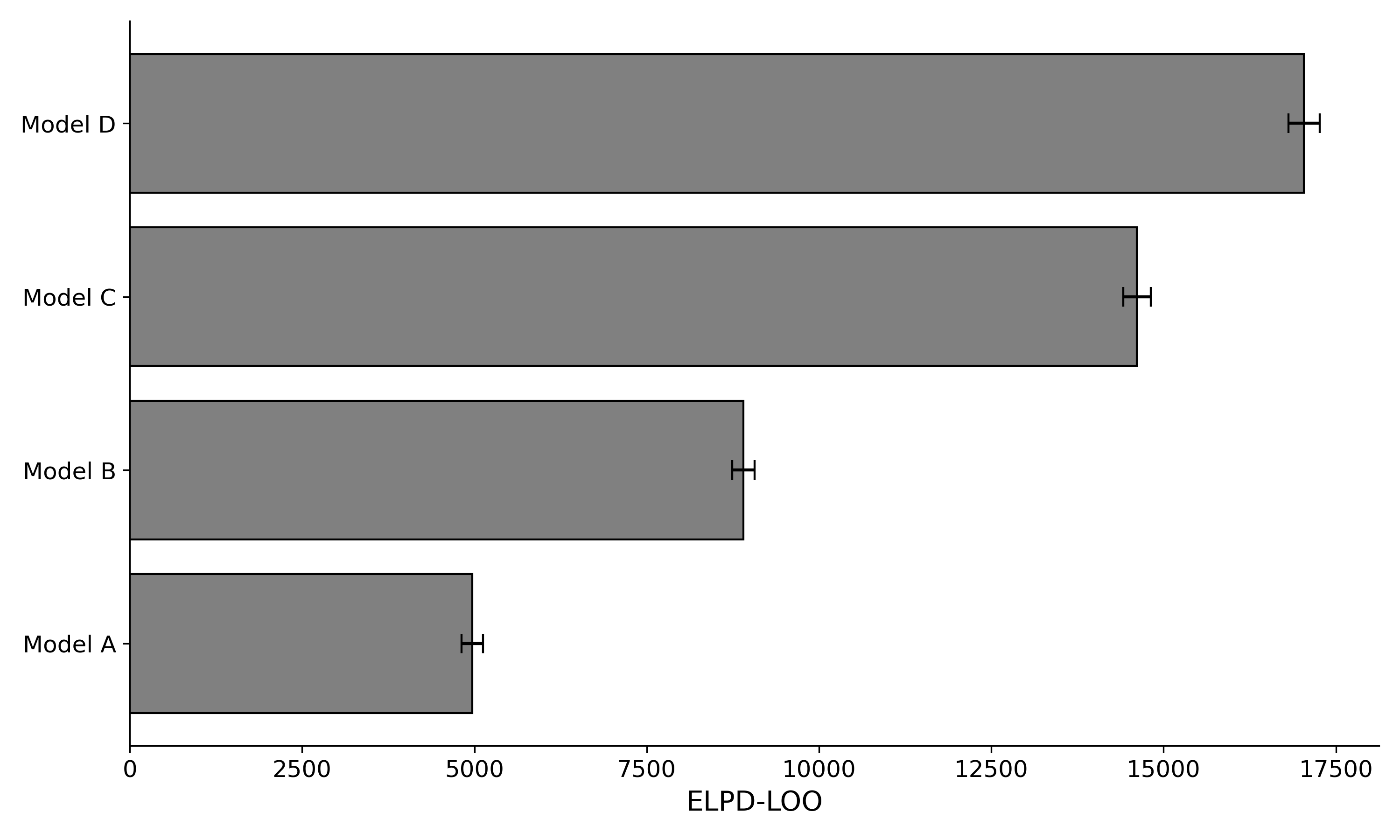}  
    \caption{Model Comparison using ELPD-LOO. Fit is across both confidence measures and change of mind data. Higher scores indicate better fit. See Methods for model specification}
    \label{fig:modelcomparison}
\end{figure}

\begin{figure}[H]
    \centering
    \includegraphics[width=1\linewidth]{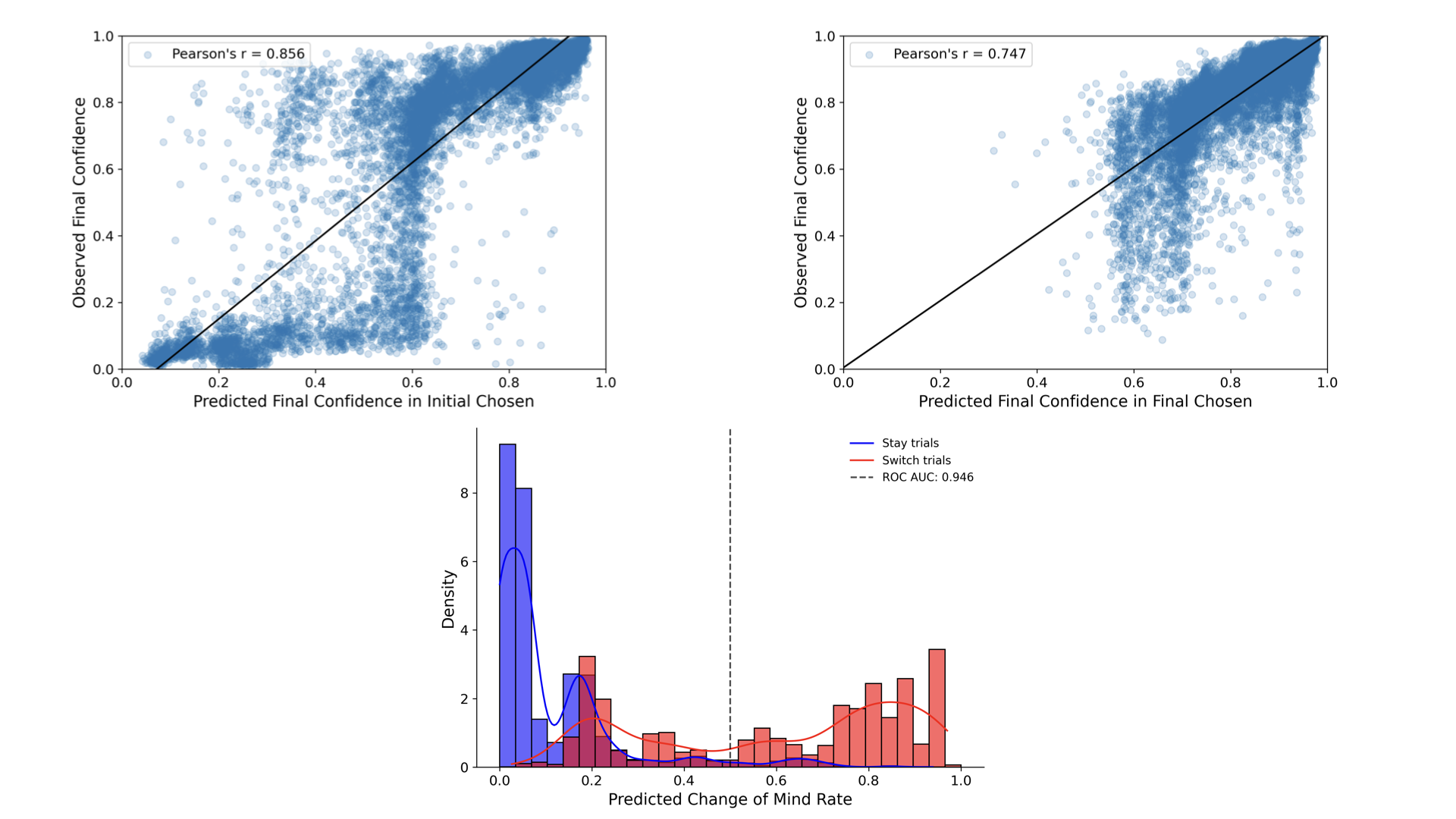}  
    \caption{Model fit on held out 10k questions from SimpleQA dataset. Top left panel: confidence in initially chosen option. Top right panel: confidence in ultimately chosen option. Bottom panel: change of mind data. Each point represents an individual trial. AUROC of change of mind data shown in legend.}
    \label{fig:heldout_modelfit_QA}
\end{figure}

We found a significant correlation between initial confidence in the initially chosen option and tendency to change mind (r=-0.11, \(p<0.0001\)). However, as evident from figure (see Fig \ref{fig:DeepSeek_composite_created}A) the model was not able to follow advice in an appropriate manner: there was actually a greater tendency to change mind in the Same Advice condition, as compared to the Opposite Advice condition (47.2\% vs 33.7\%, respectively t=-15.2, \(p<0.0001\)). Furthermore, there was no significant effect of accuracy in across conditions (\(p>0.1\)). Additionally, the decrease in confidence in the initially chosen option between first and second turns was greater in the Same Advice condition compared to the Opposite Advice condition (-0.20, -0.17, respectively; see Figure \ref{fig:DeepSeek_composite_created}B). Given the profile of findings which suggests that at least in this paradigm, this model is unable to follow advice - either in terms of change of mind or change in confidence -  we did not conduct further analyses of the model.

\begin{table}[htbp]
\centering
\renewcommand{\arraystretch}{1.4} % uniform vertical spacing adjustment
\begin{tabular}{||c c c c||} 
 \hline
 Model & ELPD-LOO & Standard Error & ELPD-DIFF \\[0.5ex]
 \hline\hline
\rule{0pt}{3ex}D  & 17263 & 222 & 0 \\[0.5ex]
 C  & 14614  & 198  & 2649 \\[0.5ex]
 B     & 8905 & 163  & 8359 \\[0.5ex]
 A     & 4969  & 158  & 12294 \\[0.5ex]
 
 \hline
\end{tabular}
\caption{Model Comparison using ELPD-LOO. Fit is across both confidence measures and change of mind data. Higher scores indicate better fit. See Methods for model specification}
\label{table:ELPD}
\end{table}

\begin{table}[h!]
\centering
\begin{filecontents*}{table_3_parameter_estimates_new.csv}
Parameter,Mean,HDI_low,HDI_high
prior,1.136,1.084,1.183
shown,1.424,1.402,1.449
intercept final chosen,0.115,0.085,0.143
Opposite Advice final chosen,1.478,1.441,1.524
Same Advice final chosen,1.17,1.122,1.215
Neutral Advice final chosen,-0.041,-1.969,1.757
\end{filecontents*}
\csvreader[
    tabular={l c c},
    table head=\toprule Parameter & Mean & 95\% HDI \\ \midrule,
    table foot=\bottomrule,
    late after line=\\,
]
{table_3_parameter_estimates_new.csv}
{Parameter=\Parameter,Mean=\Mean,HDI_low=\HDIlow,HDI_high=\HDIhigh}
{\Parameter & \Mean & [\HDIlow, \HDIhigh]}
\caption{Posterior Parameter Estimates relating to model of final confidence in the ultimately chosen option (Model D).}
\label{tab:params_final}
\end{table}

\begin{table}[h!]
\centering
\begin{filecontents*}{table_1_parameter_estimates_new.csv}
Parameter,Mean,HDI_low,HDI_high
prior,1.136,1.084,1.183
shown,1.424,1.402,1.449
intercept COM,0.759,0.662,0.845
Opposite COM,4.379,4.171,4.563
Same COM,5.333,4.376,6.164
Nothing COM,-0.018,-2.008,1.985
\end{filecontents*}
\csvreader[
    tabular={l c c},
    table head=\toprule Parameter & Mean & 95\% HDI \\ \midrule,
    table foot=\bottomrule,
    late after line=\\,
]
{table_1_parameter_estimates_new.csv}
{Parameter=\Parameter,Mean=\Mean,HDI_low=\HDIlow,HDI_high=\HDIhigh}
{\Parameter & \Mean & [\HDIlow, \HDIhigh]}
\caption{Posterior Parameter Estimates relating to change of mind model (model D)}
\label{tab:params_COM}
\end{table}

\begin{table}[h!]
\centering
\begin{filecontents*}{table_2_parameter_estimates_new.csv}
Parameter,Mean,HDI_low,HDI_high
prior,1.136,1.084,1.183
shown,1.424,1.402,1.449
intercept,-0.334,-0.373,-0.297
Answer Shown - Opposite Advice,2.824,2.76,2.889
Answer Hidden - Opposite Advice,3.126,3.036,3.216
Answer Shown - Same Advice,0.462,0.389,0.542
Answer Hidden - Same Advice,1.66,1.577,1.74
Neutral Advice,0.006,-1.817,2.014
\end{filecontents*}
\csvreader[
    tabular={l c c},
    table head=\toprule Parameter & Mean & 95\% HDI \\ \midrule,
    table foot=\bottomrule,
    late after line=\\,
]
{table_2_parameter_estimates_new.csv}
{Parameter=\Parameter,Mean=\Mean,HDI_low=\HDIlow,HDI_high=\HDIhigh}
{\Parameter & \Mean & [\HDIlow, \HDIhigh]}
\caption{Posterior Parameter Estimates relating to model of initial confidence in initially chosen option.}
\label{tab:params_initial}
\end{table}
\clearpage  % Flushes all floats and starts new page

\end{document}